\documentclass[11pt]{article}

\usepackage[T1]{fontenc}
\usepackage[utf8]{inputenc}
\usepackage{lmodern}
\usepackage[margin=1in]{geometry}
\usepackage{microtype}
\usepackage{amsmath,amssymb}
\usepackage{booktabs}
\usepackage{graphicx}
\usepackage{xcolor}
\usepackage[numbers,sort&compress]{natbib}
\usepackage[colorlinks=true,linkcolor=blue!50!black,citecolor=blue!50!black,urlcolor=blue!50!black]{hyperref}
\usepackage{url}
\usepackage{longtable}
\usepackage{tikz}
\usetikzlibrary{arrows.meta,calc,positioning,decorations.pathreplacing}
\definecolor{tjblue}{HTML}{2A78D6}
\definecolor{tjorange}{HTML}{EB6834}
\definecolor{tjink}{HTML}{0B0B0B}

\newif\iftjfinal
\tjfinaltrue
\providecommand{\tjroot}{figures}
\makeatletter
\newcommand{\tjset}[2]{\expandafter\gdef\csname tj@\tj@model @#1\endcsname{#2}}
\newcommand{\tjload}[2]{%
  \gdef\tj@model{#1}%
  \IfFileExists{\tjroot/#2/numbers.tex}{\input{\tjroot/#2/numbers.tex}}%
    {\typeout{ThinkJev: \tjroot/#2/numbers.tex not found; #1 numbers are placeholders}}%
  \IfFileExists{\tjroot/#2/example.tex}{\input{\tjroot/#2/example.tex}}{}%
  \tj@deftab{#1}{#2}{per_source}\tj@deftab{#1}{#2}{data}}
\newcommand{\tjval}[2]{\ifcsname tj@#1@#2\endcsname\csname tj@#1@#2\endcsname\else 0\fi}
\newcommand{\tj@deftab}[3]{%
  \IfFileExists{\tjroot/#2/#3.tex}%
    {\expandafter\xdef\csname tj@tab@#1@#3\endcsname{\noexpand\@@input \tjroot/#2/#3.tex }}%
    {\iftjfinal\PackageError{thinkjev}{Missing table #2/#3.tex}{Regenerate with make_figures.py}\fi
     \expandafter\gdef\csname tj@tab@#1@#3\endcsname{\multicolumn{2}{l}{\tjmissing{#1:#3.tex}} \\}}}
\newcommand{\tjtab}[2]{\csname tj@tab@#1@#2\endcsname}
\newcommand{\tjnum}[2]{%
  \@ifundefined{tj@#1@#2}%
    {\iftjfinal\PackageError{thinkjev}{Missing number #1:#2}{Regenerate numbers.tex with make_figures.py}\fi
     \tjmissing{#1:#2}}%
    {\csname tj@#1@#2\endcsname}}
\makeatother
\newcommand{\tjmissing}[1]{\textcolor{red}{\texttt{[\detokenize{#1}]}}}

\newcommand{\tjfig}[3]{
  \IfFileExists{\tjroot/#1/#2.pdf}%
    {\includegraphics[width=#3]{\tjroot/#1/#2.pdf}}%
    {\fbox{\parbox[c][0.28\textheight][c]{#3}{\centering\tjmissing{#1/#2.pdf}}}}}

\tjload{4b}{qwen3.5-4b}
\tjload{9b}{qwen3.5-9b}
\tjload{27b}{qwen3.5-27b}
\tjload{q38}{qwen3.8-27b}

\newcommand{\ie}{i.e.,\ }
\newcommand{\eg}{e.g.,\ }

\title{ThinkJev: One Thought, Every Budget\\ for Calibrated Typed Decisions}
\author{Xiaolu Ding\quad Kai Yang\quad Dongdong Liu\thanks{Corresponding author: \href{mailto:liudongdong@bosssoft.com.cn}{\texttt{liudongdong@bosssoft.com.cn}}.}\\[0.5ex]
  Fujian Bosssoft Co., Ltd., Fuzhou, China}
\date{}

\begin{document}
\maketitle

\begin{abstract}
Many uses of language models end in a typed decision: a choice among a closed set of options, whose
probabilities code can act on. Reading those probabilities
from one forward pass is fast but often wrong on decisions that take several steps, such as
arithmetic, counting, and date comparison. Letting the model think first fixes most of these
errors, but costs seconds per decision instead of milliseconds. To pay for thinking only where it
helps, ThinkJev builds a fast and a slow system from one open hybrid-thinking model that can
answer directly or after a thought. System~1 reads the model's probabilities for the option
letters when it answers directly; System~2 reads the same letters after the model has thought
on the same cached prompt, so escalating repeats no work and both systems give
probabilities over the same options. Because the letters can be read after any number of thought
tokens, one sampled thought also gives the answer under every thinking budget. Without training,
ThinkJev lets System~1 answer alone only where a statistical test on labeled dev data guarantees
that its accepted answers stay under a target error rate. At a 5\% target, Qwen3.5-27B answers
\tjnum{27b}{s1_ltt_coverage_at_5pct} of \tjnum{27b}{rows_test} test decisions this way with a
realized error of \tjnum{27b}{s1_ltt_test_risk_at_5pct} (Qwen3.5-4B and 9B:
\tjnum{4b}{s1_ltt_coverage_at_5pct} and \tjnum{9b}{s1_ltt_coverage_at_5pct}). For the rest, stopping each thought as soon as its answer is
confident reaches \tjnum{27b}{anytime_test_accuracy} accuracy, against
\tjnum{27b}{always_s2_accuracy} for always thinking, with \tjnum{27b}{anytime_thinking_share} of
the thought tokens. Choosing when to think from System~1's confidence alone saves less, because the
decisions System~1 is least sure of are often not the ones thinking fixes. On the public items of JevBench,
an independent benchmark of typed-decision services, Qwen3.5-27B thinks on
\tjnum{27b}{jb_gate_escalation} and is not distinguishable in accuracy from the two thinking APIs
listed (exploratory). Code and data generators are
available at \url{https://github.com/ding-xiaolu/ThinkJev}.
\end{abstract}

\section{Introduction}
\label{sec:intro}

Many uses of language models end in a decision that code acts on rather than a person reads:
which queue a ticket goes to, whether a message breaks a policy, whether an extracted record
satisfies a rule. We call these typed decisions: the answer is one of a fixed set of options, as
in a multiple-choice, yes/no, or rating question. For them, a probability for each option is more
useful than generated text. It needs no parsing, and its confidence can decide whether to act, ask
a person, or escalate. Commercial decision APIs return answers in this form, with a probability
per option and a confidence derived from them~\citep{typesafe2026api,typesafe2026confidence}. Open
methods such as SemIf~\citep{semif2026} and AnyJev~\citep{anyjev2026} obtain calibrated option
probabilities from any open model in one forward pass, by reading the model's next-token logits
for the option letters.

One forward pass is not enough for decisions that take several steps: date arithmetic, counting
under conditions, chained rules, stacked negation. One vendor lists exactly these as weak spots of
its decision model~\citep{typesafe2026jaggedness}. On the multi-step decisions we build
(Section~\ref{sec:setup}), a single forward pass answers only \tjnum{27b}{block_weak_s1} of them
correctly with Qwen3.5-27B (\tjnum{4b}{block_weak_s1} with Qwen3.5-4B). SemIf and AnyJev stop at
this single pass.

Letting the model think before it answers fixes most of these errors: after thinking, Qwen3.5-27B
answers \tjnum{27b}{block_weak_s2} of the same decisions correctly. Thinking is expensive, though.
On one GPU, a direct answer takes Qwen3.5-4B about \tjnum{4b}{lat_prefill_ms}~ms to process the
prompt and \tjnum{4b}{lat_s1_marginal_ms}~ms to read the answer, while thinking takes a median of
\tjnum{4b}{lat_s2_seconds}~s (Section~\ref{sec:latency}), and on routine decisions
thinking adds little accuracy (Table~\ref{tab:main}, block~A). A deployment therefore wants both:
a direct answer for the decisions it can settle, and thinking for the rest. This raises three
questions. Which decisions may skip thinking, and with what guarantee on their error rate? How
long should the rest think, when the right budget differs from one decision to the next? And can
the two be combined without running two models or processing the prompt twice?

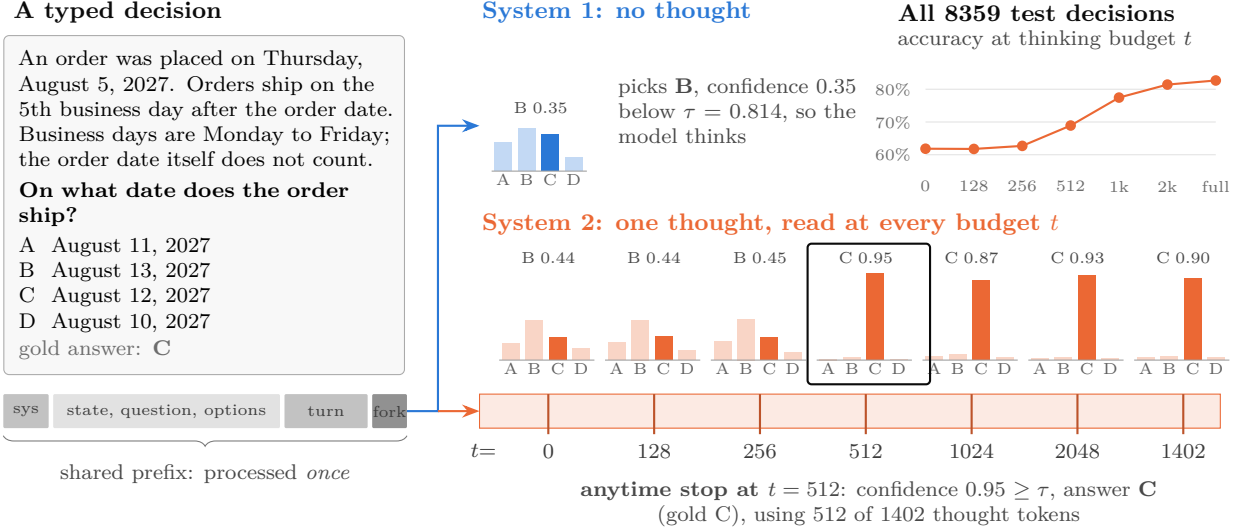
\begin{figure*}[t]
  \centering
  \def\ex#1{\tjval{27b}{ex_#1}}
\newcommand{\tjbars}[5]{%
  \edef\tjgold{\ex{gold}}%
  \foreach \L [count=\i from 0] in {A,B,C,D} {
    \pgfmathsetmacro{\h}{#5*\tjval{27b}{ex_#3_p\L}}
    \ifx\L\tjgold \def\tjc{#4}\else \def\tjc{#4!28}\fi
    \fill[\tjc] ($(#1+\i*3.1, #2)$) rectangle ++(2.3, \h);
    \node[font=\tiny, text=tjink!65, below, inner sep=0.8pt] at ($(#1+\i*3.1+1.15, #2)$) {\L};
  }
  \draw[tjink!40] ($(#1-0.4, #2)$) -- ++(12.6, 0);}

\begin{tikzpicture}[x=1mm, y=1mm, font=\footnotesize, >=Stealth,
    note/.style={font=\scriptsize, text=tjink!75, align=left},
    head/.style={font=\footnotesize\bfseries, text=tjink, anchor=west}]

  \node[head] at (0, 2) {A typed decision};
  \node[draw=tjink!35, rounded corners=2pt, fill=tjink!3, anchor=north west, text width=49mm,
        inner sep=2mm, font=\scriptsize, align=left] (card) at (0, -1)
    {\ex{state}\\[2pt] \textbf{\ex{question}}\\[2pt]
     \begin{tabular}{@{}l@{\ \ }l@{}}
       A & \ex{opt_A} \\ B & \ex{opt_B} \\ C & \ex{opt_C} \\ D & \ex{opt_D}
     \end{tabular}\\[1pt]
     {\color{tjink!60} gold answer: \textbf{\ex{gold}}}};

  \def\sy{-53}
  \foreach \x/\w/\c in {0/6/tjink!25, 6.6/30/tjink!12, 37.2/11/tjink!25, 48.8/4.6/tjink!45}
    \fill[\c, rounded corners=0.6pt] (\x, \sy) rectangle ++(\w, 4.5);
  \foreach \x/\t in {3/sys, 21.6/{state, question, options}, 42.7/turn, 51.1/fork}
    \node[font=\tiny, text=tjink!80] at (\x, \sy+2.25) {\t};
  \draw[decorate, decoration={brace, amplitude=1.4mm, mirror}, tjink!55] (0, \sy-1) -- (53.4, \sy-1);
  \node[note, anchor=north, align=center] at (26.7, \sy-3.4) {shared prefix: processed \emph{once}};

  \draw[->, thick, tjblue] (53.4, \sy+2.25) -- (57.5, \sy+2.25) |- (63, -13);
  \node[head, text=tjblue] at (62, 2) {System 1: no thought};
  \tjbars{65}{-19}{s1}{tjblue}{16}
  \node[font=\tiny, text=tjink!75] at (71.1, -10.6) {\ex{s1_pick}\;\ex{s1_conf}};
  \node[note, anchor=west, text width=33mm] at (80, -11)
    {picks \textbf{\ex{s1_pick}}, confidence \ex{s1_conf}\\
     below $\tau = \ex{tau}$, so the model thinks};

  \draw[->, thick, tjorange] (57.5, \sy+2.25) -- (63, \sy+2.25);
  \node[head, text=tjorange] at (62, -26) {System 2: one thought, read at every budget $t$};
  \fill[tjorange!14] (63, \sy) rectangle (161, \sy+4.5);
  \draw[tjorange] (63, \sy) rectangle (161, \sy+4.5);
  \foreach \n/\lab/\x in {t0/0/66, t128/128/80, t256/256/94, t512/512/108, t1024/1024/122,
                           t2048/2048/136, full/\ex{think_tokens}/150} {
    \draw[tjorange!80!black, thick] (\x+6.1, \sy-0.8) -- (\x+6.1, \sy+4.5);
    \node[font=\scriptsize, text=tjink!80, below] at (\x+6.1, \sy-0.8) {$\lab$};
    \tjbars{\x}{-44}{\n}{tjorange}{12}
    \node[font=\tiny, text=tjink!75] at (\x+6.1, -30.4) {\ex{\n_pick}\;\ex{\n_conf}};
  }
  \node[font=\scriptsize, text=tjink!80, anchor=east] at (66.6, \sy-3) {$t{=}$};
  \draw[tjink, thick, rounded corners=1.5pt] (106.4, -47.2) rectangle (122.6, -28.6);
  \node[note, anchor=north, align=center] at (114.5, \sy-5.6)
    {\textbf{anytime stop at $t = 512$}: confidence \ex{t512_conf} $\ge \tau$, answer
     \textbf{\ex{t512_pick}}\\ (gold \ex{gold}), using 512 of \ex{think_tokens} thought tokens};

  \node[head] at (117, 2) {All \tjnum{27b}{rows_test} test decisions};
  \node[note, anchor=west] at (117, -1.6) {accuracy at thinking budget $t$};
  \begin{scope}[shift={(122, -19)}]
    \foreach \n/\i in {t0/0, t128/1, t256/2, t512/3, t1024/4, t2048/5, full/6} {
      \pgfmathsetmacro{\yy}{(\tjval{27b}{curve_acc_\n}-0.55)/0.3*13}
      \coordinate (c\i) at (\i*6.4, \yy);
    }
    \foreach \v/\l in {0.6/60\%, 0.7/70\%, 0.8/80\%} {
      \pgfmathsetmacro{\yy}{(\v-0.55)/0.3*13}
      \draw[tjink!12] (0, \yy) -- (38.4, \yy);
      \node[font=\tiny, text=tjink!60, anchor=east] at (-0.6, \yy) {\l};
    }
    \draw[tjorange, thick] (c0) -- (c1) -- (c2) -- (c3) -- (c4) -- (c5) -- (c6);
    \foreach \i in {0,...,6} \fill[tjorange] (c\i) circle (0.7);
    \foreach \lab [count=\i from 0] in {0, 128, 256, 512, 1k, 2k, full}
      \node[font=\tiny, text=tjink!60, below] at (\i*6.4, 0) {\lab};
  \end{scope}
\end{tikzpicture}
  \caption{ThinkJev on one test decision, Qwen3.5-27B. Left: the part of the prompt that the two
  systems share is processed once. Top: System~1, the direct answer, gives calibrated option
  probabilities; its confidence is below the threshold $\tau$ chosen on dev, so the model thinks.
  Bottom: one sampled thought, with the same answer slots read after its first $t$ tokens (each read
  calibrated with its own temperature; checkpoints not drawn to scale). Each read is the answer
  under a thinking budget of $t$, so the policy can stop at the first confident one. Right: accuracy
  at each budget over all test decisions, with System~1 at $t = 0$. The example was chosen for
  illustration and is not typical; all values are generated from the released outputs
  (\texttt{example.tex}).}
  \label{fig:overview}
\end{figure*}

Our starting point is an observation about hybrid-thinking models such as the Qwen3
family~\citep{qwen2025qwen3}, which answer either directly or after a thought, depending on how
the chat template is set. For the same decision, the direct prompt and the thinking prompt are
identical up to the thinking block, and in both the answer can be read at the same place: the
model's next-token logits for the option letters A, B, \dots\ at the answer position, which we
call the answer slots. One model therefore provides two systems on one prompt. System~1 is the
direct answer: the answer slots read after an empty thought. System~2 is the answer after
thinking: the same slots read after a sampled thought. If the thought is cut after its first $t$
tokens and closed with the model's own early-stop sentence, reading the slots gives exactly the
answer the model would have given under a thinking budget of $t$ tokens. One sampled thought
therefore gives a calibrated answer at every budget, with System~1 at budget zero
(Figure~\ref{fig:overview}).

ThinkJev\footnote{The name refers to the typed-decision request format of TypeSafe's Jev, which
ThinkJev's server accepts (Appendix~\ref{app:server}). ThinkJev is independent of TypeSafe, and Jev
is not evaluated in this paper.} builds a cascade on this observation. It needs no training and
answers the three questions in turn. System~1 answers on its own only when its confidence reaches
a threshold. The threshold is chosen on a labeled development (dev) split with a statistical test,
Learn-then-Test, which guarantees that the error rate of the answers System~1 gives stays under a
target the user sets. Every other decision is escalated: the model thinks, continuing from the
prompt it has already processed, and the answer is read from the same option letters, again as
calibrated probabilities. The user can fix the thinking budget, or let the model stop as soon as
its answer is confident. No generated text is parsed. We make the following contributions; each
is tied to a measurement in Section~\ref{sec:results}.

\begin{enumerate}
  \item \textbf{Direct answers with an error guarantee} (Section~\ref{sec:risk}). System~1 gives
  calibrated option probabilities from one forward pass. A threshold on its confidence, chosen by
  Learn-then-Test~\citep{angelopoulos2021ltt} on dev, bounds the error rate of the decisions it
  answers on its own; decisions generated from one template count as one sample, so near-duplicates
  do not inflate the guarantee. On every Qwen3.5 model the error rate on test stays under each
  target for which the threshold accepts any decision. How many decisions System~1 may answer
  depends on the model: at the strictest targets the smaller models accept none. On Qwen3.8-27B
  the test error rate slightly exceeds the loosest target (Figure~\ref{fig:risk},
  Table~\ref{tab:risk}).
  \item \textbf{An answer at every thinking budget} (Section~\ref{sec:anytime}). Cutting one sampled
  thought after $t$ tokens and reading the answer slots gives the answer under a budget of $t$, so
  one run gives the whole curve of accuracy and calibration against the budget, with System~1 at
  budget zero. An anytime policy can then stop thinking at the first checkpoint where the answer is
  confident (Figure~\ref{fig:anytime}).
  \item \textbf{One model as its own cascade, and the limit of gating on confidence}
  (Sections~\ref{sec:method} and~\ref{sec:h1}). System~1 and System~2 come from the same model and
  the same processed prompt and read the same option letters, so the cascade needs no second model,
  no trained weights, and no parsing. Deciding when to think from System~1's confidence sends
  thinking to the decisions where System~1 is weak, but saves little thinking overall. We trace
  this to a difference between two targets: predicting where System~1 is wrong is not the same as
  predicting where thinking fixes the answer (Figure~\ref{fig:pareto}, Table~\ref{tab:main},
  Section~\ref{sec:where}).
  \item \textbf{A generated benchmark of multi-step decisions} (Section~\ref{sec:setup}). Eight
  families of decisions (date arithmetic, conditional counting, chained rules, stacked negation,
  multi-step arithmetic, logic grids, graph reachability, and scheduling), each at four difficulty
  levels, written natively in English and in Chinese, plus chess puzzles. The decisions are
  generated by code and every label is computed by code (Figures~\ref{fig:tasks}
  and~\ref{fig:difficulty}).
  \item \textbf{An open server} (Appendix~\ref{app:server}) for typed decisions that accepts the
  requests and returns the responses of a public decision API (TypeSafe's
  \texttt{POST /v1/systemone}), so that the API's own SDK can call a local ThinkJev model.
\end{enumerate}

The text and figures use Qwen3.5-27B, the most accurate of the four models we run. Every quantity
is also reported for the other three models (Section~\ref{sec:setup}), and every criterion that
fails on a model is reported as failed.

\section{Background and related work}
\label{sec:related}

Deciding per query whether to think, and how much, is an active topic. Table~\ref{tab:related}
places ThinkJev among the closest lines of work; the paragraphs below say what each line does and
what is different here.

\paragraph{Typed-decision models.}
TypeSafe's Jev, a commercial decision model, answers choice, yes/no, and score questions with
probabilities and a confidence computed from them~\citep{typesafe2026api,typesafe2026confidence},
and TypeSafe's own documentation recommends gating actions on that confidence, for example routing
low-confidence decisions to a person~\citep{typesafe2026routing}. SemIf~\citep{semif2026} reproduces
the readout on open models by reading the logits of single-token option letters after a fixed direct
prompt; ThinkJev reuses its prompt and code unchanged. AnyJev~\citep{anyjev2026} adds
cyclic-shift debiasing, batch priors, coverage at a target risk, and hidden-state heads. Both are
System~1 only. In concurrent work, OpenJev~\citep{openjev2026} lets a request set a thought budget
in advance, after which the answers are read; ThinkJev reads every budget from one thought and
decides per decision whether to think. ThinkJev's gate is TypeSafe's confidence-gated routing
pattern with the low-confidence branch sent to the same model's thinking mode, which answers through
the same slots.

\paragraph{Option readout and calibration.}
Language models prefer some option positions over others~\citep{zheng2024selectors}; averaging
the read over cyclic shifts of the option list removes any position bias that is additive in
logit space. Temperature scaling~\citep{guo2017calibration} fit on a dev split calibrates each
readout; we fit one temperature per readout (System~1, each anytime checkpoint, and the final
System~2).

\paragraph{Adaptive thinking.}
Training-based methods change the weights so that the model itself allocates reasoning:
When2Think uses RLVR post-training with a difficulty-aware efficiency
bonus~\citep{shim2026when2think}, AdapThink uses RL post-training with adaptive thinking
preferences~\citep{xu2026adapthink}, and CAT folds the model's self-certainty into preference
optimization~\citep{jiang2026cat}. ThinkJev trains nothing. Sonata, from Adaptive
Thinking~\citep{li2026adaptive}, trains a light adapter on prefill hidden states to predict
self-consistency and sets the thinking budget before any thought; our probe
router belongs to this family, reading the same model's hidden states at the last position of the
shared prefix. DART~\citep{lee2026dart} is training-free and closest in spirit: it samples two no-think
drafts, answers directly when they agree, and otherwise predicts a budget from draft entropy, and
it applies to free-form answers such as math and code. For closed-set decisions, ThinkJev replaces
the sampled drafts with one forward pass that already yields the full calibrated option
distribution. DEER~\citep{yang2025deer} exits a reasoning trace early when a trial answer is
confident; with fixed option slots, a trial answer is one short forward pass that appends the
early-stop sentence, and the anytime curve starts at System~1. Studies of hybrid models report
that reasoning leaks into the no-think mode~\citep{wang2025demystifying} and compare mode-selection
strategies empirically~\citep{tan2025zerostep,ning2026hrbench}. In our System~1 read, nothing is
generated: the slots are read directly after the closed empty thinking block.

\paragraph{Cascades, routing, and risk control.}
FrugalGPT~\citep{chen2023frugalgpt} cascades across models with a learned answer scorer, and
RouteLLM~\citep{ong2024routellm} trains a router between a strong and a weak model on preference
data. ThinkJev cascades within one model on one cached prefix, so escalation reuses the prefill
and System~2's answer is calibrated on the same scale. Learn-then-Test~\citep{angelopoulos2021ltt}
turns a threshold choice into a multiple-testing problem with a finite-sample guarantee; we apply
it to the System~1 acceptance threshold.

\begin{table}[t]
  \centering
  \small
  \begin{tabular}{@{}p{0.2\linewidth}p{0.36\linewidth}p{0.36\linewidth}@{}}
    \toprule
    Work & What it does & What ThinkJev adds or changes \\
    \midrule
    Jev~\citep{typesafe2026api} & Closed decision model with calibrated probabilities &
      Open weights, no training, any hybrid-thinking model; a System~2 on the same answer slots \\
    SemIf, AnyJev~\citep{semif2026,anyjev2026} & System~1 readout from any LLM; debiasing,
      temperatures, hidden-state heads & System~2 on the shared prefix, gating, anytime readout,
      risk guarantee, typed-decision server \\
    Adaptive Thinking~\citep{li2026adaptive} & Adapter on prefill hidden states sets the budget
      before thinking & Not run; our probe router, of the same family, is compared
      with the gate (Section~\ref{sec:h1}) \\
    DART~\citep{lee2026dart} & Training-free; agreement of two sampled no-think drafts &
      One forward gives the calibrated option distribution; closed-set decisions only \\
    DEER~\citep{yang2025deer} & Early exit when a trial answer is confident & Trial answer is
      a slot read; the curve's zero point is System~1 \\
    When2Think, AdapThink, CAT~\citep{shim2026when2think,xu2026adapthink,jiang2026cat} &
      Post-training so the model allocates its own reasoning & No weight updates \\
    Zero-Step Thinking, HRBench~\citep{tan2025zerostep,ning2026hrbench} & Empirical studies and a
      benchmark of mode selection & Typed decisions with calibrated probabilities at both ends and
      a risk guarantee \\
    \bottomrule
  \end{tabular}
  \caption{Closest related work. Descriptions follow each work's own abstract or documentation.}
  \label{tab:related}
\end{table}

\section{Method}
\label{sec:method}

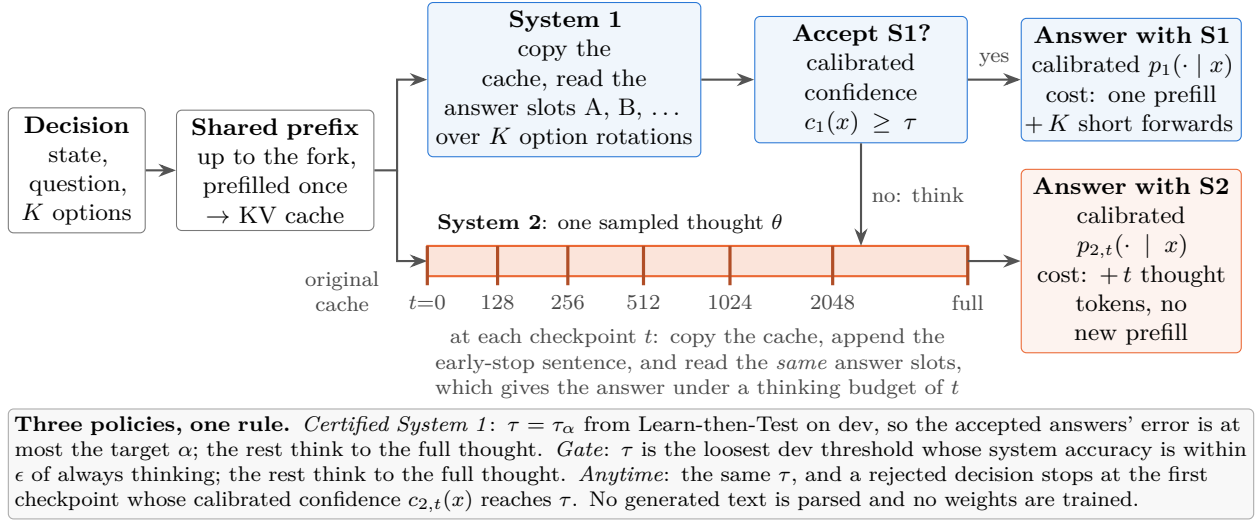
\begin{figure}[t]
  \centering
  \begin{tikzpicture}[
    font=\footnotesize, >=Stealth,
    box/.style={draw=tjink!55, rounded corners=2pt, align=center, inner sep=3pt, fill=white,
                minimum height=15mm},
    sone/.style={box, draw=tjblue, fill=tjblue!7},
    stwo/.style={box, draw=tjorange, fill=tjorange!8},
    flow/.style={->, draw=tjink!70, thick},
    note/.style={font=\scriptsize, text=tjink!75, align=center},
    cost/.style={font=\scriptsize\itshape, text=tjink!60, align=left}]

  \node[box, text width=16mm] (row) at (0,0) {\textbf{Decision}\\ state,\\ question,\\ $K$ options};
  \node[box, text width=24mm, right=4mm of row] (prefix)
    {\textbf{Shared prefix}\\ up to the fork,\\ prefilled once\\ $\to$ KV cache};
  \draw[flow] (row) -- (prefix);
  \coordinate (split) at ($(prefix.east)+(3mm,0)$);

  \node[sone, text width=34mm, anchor=west] (s1) at ($(prefix.east)+(7mm,12mm)$)
    {\textbf{System 1}\\ copy the cache, read the\\ answer slots A, B, \dots\\ over $K$ option rotations};
  \node[sone, text width=26mm, right=7mm of s1] (gate)
    {\textbf{Accept S1?}\\ calibrated\\ confidence\\ $c_1(x) \ge \tau$};
  \node[sone, text width=27mm, right=7mm of gate] (ans1)
    {\textbf{Answer with S1}\\ calibrated $p_1(\cdot \mid x)$\\ cost: one prefill\\ $+\, K$ short forwards};
  \draw[flow] (prefix.east) -- (split) |- (s1.west);
  \draw[flow] (s1) -- (gate);
  \draw[flow] (gate) -- node[above=1pt, note] {yes} (ans1);

  \coordinate (t0) at ($(prefix.east)+(7mm,-12mm)$);
  \coordinate (tn) at ($(gate.east |- t0)+(0,0)$);
  \draw[tjorange, thick, fill=tjorange!18] ($(t0)+(0,-2mm)$) rectangle ($(tn)+(0,2mm)$);
  \node[note, anchor=south west, text=tjink] at ($(t0)+(0,2.4mm)$)
    {\textbf{System 2}: one sampled thought $\theta$};
  \draw[flow] (split) |- ($(t0)+(0,0)$);
  \node[note, anchor=north east] at ($(split |- t0)+(-1mm,0)$) {original\\ cache};
  \foreach \x/\t in {0/{t{=}0}, 0.13/128, 0.26/256, 0.40/512, 0.56/1024, 0.75/2048, 1.0/\text{full}} {
    \draw[tjorange!75!black, very thick] ($($(t0)!\x!(tn)$)+(0,-3.2mm)$) -- ($($(t0)!\x!(tn)$)+(0,2mm)$);
    \node[note, below] at ($($(t0)!\x!(tn)$)+(0,-3.2mm)$) {$\t$};
  }
  \node[note, below, text width=95mm] at ($($(t0)!0.5!(tn)$)+(0,-7.6mm)$)
    {at each checkpoint $t$: copy the cache, append the early-stop sentence, and read the
     \emph{same} answer slots, which gives the answer under a thinking budget of $t$};

  \draw[flow] (gate.south) -- node[right, note] {no: think} ($(gate.south |- t0)+(0,2mm)$);
  \node[stwo, text width=27mm, anchor=west] (ans2) at ($(ans1.west |- t0)$)
    {\textbf{Answer with S2}\\ calibrated $p_{2,t}(\cdot \mid x)$\\ cost: $+\,t$ thought\\ tokens, no new prefill};
  \draw[flow] (tn) -- (ans2);

  \node[draw=tjink!35, rounded corners=2pt, fill=tjink!3, inner sep=2.5pt, anchor=north west,
        text width=\dimexpr\textwidth-6pt\relax, font=\scriptsize, align=left]
    at ($(row.west |- t0)+(0,-19.5mm)$)
    {\textbf{Three policies, one rule.} \textit{Certified System~1}: $\tau = \tau_\alpha$ from
     Learn-then-Test on dev, so the accepted answers' error is at most the target $\alpha$; the rest
     think to the full thought. \textit{Gate}: $\tau$ is the loosest dev threshold whose system
     accuracy is within $\epsilon$ of always thinking; the rest think to the full thought.
     \textit{Anytime}: the same $\tau$, and a rejected decision stops at the first checkpoint whose
     calibrated confidence $c_{2,t}(x)$ reaches $\tau$. No generated text is parsed and no weights
     are trained.};
\end{tikzpicture}
  \caption{ThinkJev on one decision: the same answer slots read twice after one processed prompt
  prefix. System~1 is read on a copy of the key-value cache; if its calibrated confidence reaches
  $\tau$, it answers. Otherwise the model thinks, continuing on the original cache, and the same
  slots are read after the first $t$ thought tokens (checkpoints not to scale). The three policies
  of Sections~\ref{sec:gate} and~\ref{sec:risk-method} (direct answers under an error guarantee, the
  gate, and the anytime policy) differ only in how $\tau$ is chosen and where an escalated decision
  stops thinking.}
  \label{fig:method}
\end{figure}

A decision $x$ consists of a state (the facts to decide on), a question, and $K$ options
($2 \le K \le 16$), each with an ID and a description. The model is a hybrid-thinking language
model, whose chat template selects either a direct answer or an answer after a thought. Figure~\ref{fig:method} shows the pipeline.
Three requirements shape it. Both systems must return probabilities over the same options, so that
either can answer and code can act on the result the same way. Escalating to System~2 must not redo the work
already spent on System~1. And the user, not the calibration, should decide how long a decision
may think. Reading the same answer slots after one shared prompt prefix (Sections~\ref{sec:s1}
to~\ref{sec:s2}) meets the first two; reading them at checkpoints inside one thought
(Section~\ref{sec:anytime-method}) meets the third.

\subsection{System 1: reading the answer slots}
\label{sec:s1}

System~1 is SemIf's \texttt{direct-options-v1} readout~\citep{semif2026}, whose code we copy
unchanged. The
prompt lists the options under the letters A, B, \dots, and is rendered with thinking disabled,
which for Qwen3.5 closes an empty thinking block before the answer position. Each letter must be
exactly one token, and appending it must not change the tokenization of the prompt. One forward
pass gives the logits $z \in \mathbb{R}^K$ of the $K$ letter tokens at the answer position, and
\begin{equation}
  p_1(k \mid x) = \operatorname{softmax}(z / T_1)_k, \qquad
  c_1(x) = \max_k p_1(k \mid x),
\end{equation}
with the temperature $T_1$ fit on dev by negative log-likelihood~\citep{guo2017calibration}.
Because SemIf's prompt is used byte for byte, the hashes of our System~1 prompts equal those of
SemIf's published direct predictions (Appendix~\ref{app:repro}).

\paragraph{Option rotations.}
Models favor some option positions~\citep{zheng2024selectors}, so a single read's confidence partly
reflects where an option is listed rather than how sure the model is, which makes it a poorer
signal for deciding when to think. Following AnyJev, we also read the prompt with the option list
cyclically shifted $K-1$ more times. The letters stay in place and each option moves with its
description, so every option appears under every letter once. With $\ell^{(j)}_k$ the log-probability of option $k$ under shift $j$, the debiased read is
\begin{equation}
  \bar p_1(k \mid x) \propto \exp\Bigl(\tfrac{1}{K} \textstyle\sum_{j=0}^{K-1} \ell^{(j)}_k\Bigr),
\end{equation}
a renormalized geometric mean that removes any position bias additive in logit space. It has its
own dev temperature. The shifted prompts share every token up to the first option description, so
the extra cost is $K-1$ short forward passes over the rest of the prompt. The gate we report
(\texttt{confidence\_rot} in the released code) uses this read, and the decisions it accepts are
answered with its most probable option.

\subsection{Shared prefix and fork}
\label{sec:fork}

An escalated decision should not pay for its prompt twice. The two systems therefore share one
pass over the prompt (the prefill, which fills the model's key-value cache), which requires
finding where their prompts diverge.

The System~1 prompt and the thinking prompt for the same decision are tokenized separately, and
the fork is the end of their longest common token prefix. It covers the system and user turns and ends inside the
assistant's opening; for Qwen3.5 the text boundary after \texttt{<think>} is not a token boundary,
so the fork is located on tokens rather than text. The shared prefix is prefilled once. Qwen3.5 mixes
full-attention layers with Gated DeltaNet layers~\citep{yang2025gateddelta}, whose recurrent state
cannot be rolled back to an earlier position. System~1 is therefore read on a copy of the cache,
and System~2 continues on the original, so escalation never repeats the prefill.

\subsection{System 2: the same slots after a thought}
\label{sec:s2}

System~2 renders the same messages with thinking enabled and samples a thought with Qwen's
recommended thinking settings (temperature 0.6, top-$p$ 0.95, top-$k$ 20) and a seed
derived from the decision's ID. If the thought closes by itself, we append \texttt{</think>} and the
tokens that follow the thinking block in the System~1 prompt. If it reaches the thinking budget $B$,
we append Qwen's early-stop sentence, which ends with \texttt{</think>}. In either case we read the
logits of the same letter tokens at the same answer position, giving $p_2(k \mid x) =
\operatorname{softmax}(z_2 / T_2)_k$ with its own temperature $T_2$, also fit on dev. No generated text is
parsed, so System~2 has no invalid outputs, and its answer has the same form as System~1's: a
probability over the same options, which the caller can use in place of System~1's without any
conversion.

\paragraph{Batch System 2.}
For thousands of decisions, thoughts are sampled with vLLM from the exact token IDs of the
thinking prompt, with the same settings and seeds, and the answer slots are then read in one
forward pass over prompt, thought, and closing tokens. On a replayed thought, this read reproduces
the System~2 logits of the single-process implementation that the server uses
(Appendix~\ref{app:repro}). Thoughts from vLLM are a different sample from the same distribution,
so latency is measured with the single-process implementation only.

\subsection{Anytime readout}
\label{sec:anytime-method}

The thinking budget a decision needs is not known in advance, and a caller may need to answer by a
deadline. Sampling one thought per budget would multiply the cost; instead, one thought is read at
several points along the way.

Let $\theta = (\theta_1, \dots, \theta_n)$ be the sampled thought, $n \le B$. For a checkpoint
$t < n$, \emph{System~2 at $t$} appends the early-stop sentence to $\theta_{1:t}$ and reads the
same slots; for $t \ge n$ it is the final System~2 read. Under a thinking budget of $t$, the same
sampler with the same seed emits the same first $t$ tokens and is then closed with the same
sentence, so System~2 at $t$ is the answer under budget $t$ for this sampled thought
(Appendix~\ref{app:anytime} states the conditions). One run therefore gives the answer at every
budget up to the full thought. In practice the thought is fed in chunks on one cache, and at each
checkpoint a copy of the cache receives the early-stop sentence and the slots are read. We use $t \in \{0, 128, 256, 512, 1024,
2048\}$ and the full thought ($\le 4096$). Each checkpoint has its own temperature $T_{2,t}$, fit on dev.

At $t = 0$ the read is the thinking template with an empty thought closed by the early-stop
sentence. It differs from System~1 only by that sentence inside the thinking block, which is why we
treat System~1 as the curve's point at budget zero; Section~\ref{sec:anytime} measures how close the
two reads are.

\subsection{Calibrated gating and the anytime policy}
\label{sec:gate}

\paragraph{Gate.} The gate thinks only where System~1 is unsure. Accept System~1 if
$c_1(x) \ge \tau$; otherwise think and answer with System~2.
The threshold $\tau$ is chosen on dev: among the thresholds whose accuracy on dev is at most
$\epsilon$ below that of always thinking, take the one with the fewest thought tokens per decision.
Our main setting, the operating point used throughout, is $\epsilon = 1$ point;
Table~\ref{tab:tolerance} also reports $\epsilon \in \{0, 2\}$. Test decisions never enter any fit
or selection.

\paragraph{Anytime policy.} The anytime policy also stops thinking once the answer is
confident. Read System~1; if $c_1(x) < \tau$, read System~2 at successive
checkpoints and stop at the first whose calibrated confidence $\max_k
\operatorname{softmax}(z_{2,t}/T_{2,t})_k$ reaches $\tau$, else use the full thought. Its cost is
the thought tokens up to the checkpoint where it stops; the short forward passes that read earlier
checkpoints are not counted (Section~\ref{sec:limits}). $\tau$ is chosen on dev by the same rule as the gate.

\subsection{Risk guarantee for System 1}
\label{sec:risk-method}

Calibration makes System~1's confidence right on average, but an application that lets System~1
act alone needs a statement about the answers it accepts: of those, at most a fraction $\alpha$ is
wrong. Learn-then-Test~\citep{angelopoulos2021ltt} gives such a statement without training: it
picks the threshold on dev so that, with probability at least $1 - \delta$ over the draw of the dev
data, the error rate of the answers System~1 gives on new data from the same distribution is at
most $\alpha$. For a target error rate (risk) $\alpha \in \{1\%, 2\%, 5\%, 10\%\}$ we choose the System~1
acceptance threshold this way at $\delta = 0.05$. Candidate thresholds are
tested on dev from strictest to loosest with fixed-sequence testing, which needs no multiplicity
correction, and testing stops at the first failure. At each candidate, the error rate of the decisions it accepts
is tested with a one-sided Clopper--Pearson upper bound. Decisions from the same group (\eg variants
of one template) are not independent, so the sample size is the number of distinct groups accepted
and the error count is scaled to it, rounded up. The sequence starts at the first threshold whose
group count would let zero errors pass, $n \ge \log\delta / \log(1-\alpha)$; this depends only on
counts, so the guarantee is unaffected. The dev threshold is applied unchanged to test, where we
report its coverage (the share of decisions System~1 answers) and its realized risk (the error
rate of those answers).

\subsection{A server for typed decisions}
\label{sec:server-method}

\texttt{thinkjev-serve} answers \texttt{POST /v1/systemone} in the request and response shapes of
TypeSafe's public API~\citep{typesafe2026api}. Each choice, yes/no, or score question becomes one
decision, runs through the gated cascade, and is answered with the calibrated probabilities of whichever
system answered; temperatures and the threshold are read from an evaluation report.
Appendix~\ref{app:server} gives the mapping, an example, and the differences from Jev.

\section{Experimental setup}
\label{sec:setup}

\paragraph{Models.} Four hybrid-thinking models, all in BF16 and pinned by exact revision
(Appendix~\ref{app:repro}): Qwen3.5-4B~\citep{qwen2026qwen35_4b}, Qwen3.5-9B,
Qwen3.5-27B~\citep{qwen2026qwen35_27b}, and Qwen3.8-27B, all run on the same decisions with the same
settings and analysis code. Qwen3.8's chat template inserts a
reasoning-effort note before the user turn in thinking mode, so its thinking prompt is rendered with
\texttt{reasoning\_effort="medium"}, the one setting that adds no note; only then do its two prompts
share the decision's prefix. Thinking uses a budget of $B = 4096$ tokens, one sampled thought per
decision (seed index 0), and the settings of Section~\ref{sec:s2}. The text and figures use Qwen3.5-27B,
the most accurate of the four; every quantity is reported for every model (Tables~\ref{tab:scale},
\ref{tab:risk}, and~\ref{tab:h1}), and the repository holds the same figures for every model.

\paragraph{Data.} The decisions come in three blocks (Table~\ref{tab:data}). Block~A holds
routine decisions that a direct answer often settles: intent routing over
BANKING77~\citep{casanueva2020banking77} and CLINC150~\citep{larson2019clinc}, detecting whether a
request should be escalated to a person (as AnyJev defines the task), natural language inference
(NLI) from WANLI~\citep{liu2022wanli}, the hand-written examples that ship with SemIf, a public set
of labeled judgment questions that SemIf also uses (``Every judgments''), and the typed-decisions
dataset~\citep{typeddecisions}. Block~B holds public reasoning
benchmarks with fixed options, which let readers compare the two systems on established tasks
(Section~\ref{sec:public}): the multiple-choice tasks of BIG-Bench Hard~\citep{suzgun2022bbh},
MMLU-Pro~\citep{wang2024mmlupro}, GPQA Diamond~\citep{rein2023gpqa}, SuperGPQA~\citep{mapteam2025supergpqa},
and MuSR~\citep{sprague2024musr}. Block~C is ours and holds multi-step decisions: the eight generated
families of Section~\ref{sec:intro} at four difficulty levels each, written in English and in
Chinese; an earlier set of 200 hard generated decisions (\texttt{hard200}); and chess puzzles from
the Lichess database~\citep{lichesspuzzles} (pick the best of four legal moves, with difficulty set
by the puzzle's rating). Only block~C has Chinese decisions. Appendix~\ref{app:data} describes the
generators. All decisions are split into dev (40\%) and test by a hash of their source and group,
so variants of one template never fall on both sides of the split. The repository stores only
decision IDs; third-party text is rebuilt locally from upstream versions pinned by commit.

\begin{table}[t]
  \centering
  \small
  \begin{tabular}{@{}llrrll@{}}
    \toprule
    Block & Source & Dev & Test & Language & License \\
    \midrule
    \tjtab{27b}{data}
    \bottomrule
  \end{tabular}
  \caption{Decisions per source and split, from the released split file
  (\texttt{benchmarks/\allowbreak manifests/\allowbreak release-split.jsonl}); licenses as stated upstream
  (\texttt{THIRD\_PARTY.md}). Block~A: routine decisions; B: public reasoning; C: generated
  multi-step decisions and chess. Every source is used by all four models.}
  \label{tab:data}
\end{table}

\paragraph{Criteria.} We fixed two success criteria and two comparisons before the runs. The
\emph{savings criterion}: at the threshold chosen on dev, the gate's test accuracy is within 1.5
points of always thinking while it escalates at most half of the decisions. The \emph{risk
criterion}: the test error rate of the decisions that the Learn-then-Test threshold lets System~1
answer stays within each target. The anytime curve is reported descriptively, without a criterion.
The \emph{ranking comparison} sets the gate against pre-routers (below). A planned comparison with
Jev on our decisions was dropped before any run, because we do not call Jev's API
(Section~\ref{sec:limits}). Each criterion is reported for every model, and a criterion that fails
is reported as failed. Temperatures, probes, and thresholds are fit on dev, and operating points
are chosen on dev; test decisions never enter any fit or selection.

\paragraph{Policies and baselines.} A policy decides, for each decision, whether System~1
answers or the decision is escalated, \ie sent to System~2. All policies are simulated offline
from one run that reads both System~1 and System~2 on every decision. The two endpoints are always
answering directly (always-S1) and always thinking (always-S2). The gate thresholds System~1's
calibrated confidence, read either once (\texttt{confidence} in the released code) or averaged
over option rotations (\texttt{confidence\_rot}, the gate we report); the anytime policy is
described in Section~\ref{sec:gate}. Pre-routers are the usual alternative to gating on an
answer's confidence: they make the same choice from the decision's text alone, before any answer
is read. We compare three. The \emph{probe router} (\texttt{router\_probe}) is a ridge regression
on the same model's hidden states at the last position of the shared prefix (layer and penalty
chosen by group cross-validation on dev). The \emph{zero-shot router} (\texttt{router\_zero\_shot})
asks the same model, with the System~1 prompt, whether the decision can be made at a glance
(Appendix~\ref{app:prompts}). The \emph{$n$-gram router} (\texttt{router\_supervised}) is a
logistic regression on hashed $n$-grams of the decision's text, trained on dev to predict whether
System~1 is right. The oracle escalates exactly the decisions where System~1 is wrong and System~2
is right.

\paragraph{Metrics.} Accuracy; mean thought tokens per decision, counting zero for decisions
System~1 answers; expected calibration error (ECE: top label, 10 equal-width bins); coverage and
realized risk at each target (Section~\ref{sec:risk-method}); and, for the ranking comparison, the
area under the test accuracy--coverage curve, normalized so that random escalation scores 0 and
the oracle 1: $(\text{policy} - \text{random}) / (\text{oracle} - \text{random})$. Differences
from the gate get 95\% confidence intervals from a bootstrap over groups (1{,}000 resamples).
Realized risk is shown with an exact binomial 95\% interval over the accepted test decisions; it
treats decisions as independent, so it is narrower than an interval over groups. We also count the
test decisions where System~1 and System~2 differ in correctness, which bounds what any gate can
gain (Section~\ref{sec:limits}).

\paragraph{Compute and latency.} Thoughts are sampled with vLLM across GPUs; answer slots are
read in a separate process with one GPU each. Latency is measured for Qwen3.5-4B with the
single-process implementation, on one H200 at batch size 1, over a stratified subset of
\tjnum{4b}{lat_rows} test decisions, with nothing else running on the machine.

\section{Results}
\label{sec:results}

All numbers below come from the evaluation reports of complete runs; each table and figure names
the report fields it reads. Thresholds were chosen on dev and applied unchanged to test. The
text uses Qwen3.5-27B (Section~\ref{sec:setup}); Section~\ref{sec:scale} and
Tables~\ref{tab:risk} and~\ref{tab:h1} give the same quantities for every model.

\subsection{Main result}
\label{sec:main}

On \tjnum{27b}{rows_test} test decisions, Qwen3.5-27B answers \tjnum{27b}{always_s1_accuracy}
correctly when it always answers directly and \tjnum{27b}{always_s2_accuracy} when it always
thinks, spending \tjnum{27b}{always_s2_mean_think_tokens} thought tokens per decision on average.
With the threshold chosen on dev (at most one point below always thinking on dev), the gate reaches
\tjnum{27b}{gate_test_accuracy} with \tjnum{27b}{gate_mean_think_tokens} thought tokens per
decision, \tjnum{27b}{gate_thinking_share} of what always thinking spends; the anytime policy
reaches \tjnum{27b}{anytime_test_accuracy} with \tjnum{27b}{anytime_mean_think_tokens}
(\tjnum{27b}{anytime_thinking_share}).

\begin{figure}[t]
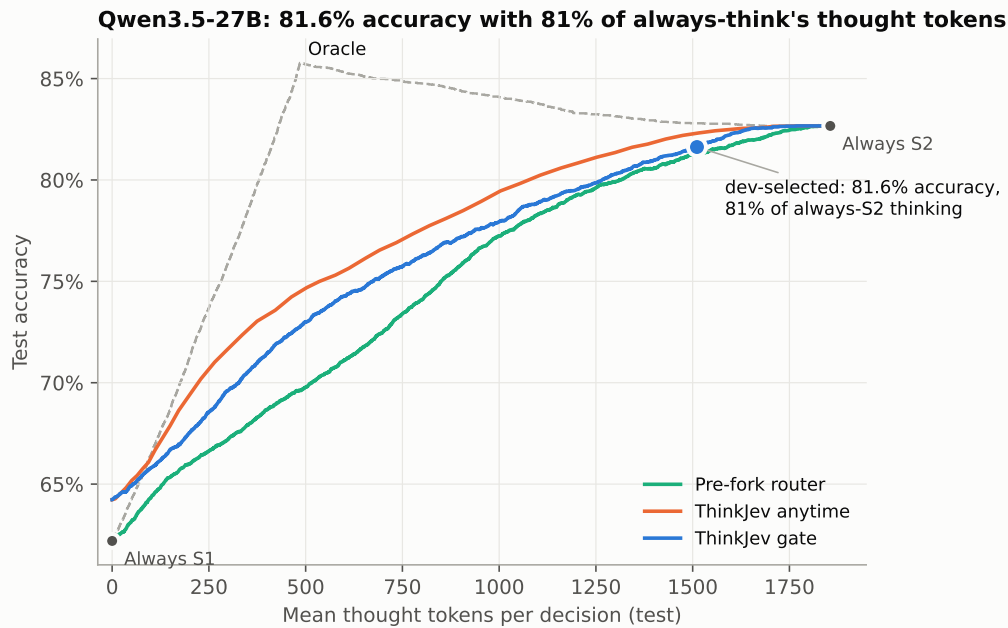

  \centering
  \tjfig{qwen3.5-27b}{pareto}{0.82\linewidth}
  \caption{Test accuracy against thought tokens per decision for Qwen3.5-27B. Each curve sweeps the
  threshold of one policy: the gate, the anytime policy, and the probe router. The oracle escalates
  exactly the decisions that System~1 gets wrong and System~2 gets right. Always answering
  directly (always-S1) and always thinking (always-S2) are the endpoints; the marked point is the
  gate's threshold chosen on dev and applied unchanged to test. Source fields:
  \texttt{cost\_curves\_test}, \texttt{anytime\_test.curves},
  \texttt{anytime\_test.dev\_selected\_operating\_points}.}
  \label{fig:pareto}
\end{figure}

\textbf{The savings criterion fails on every model.} Its accuracy half holds: the gate is
\tjnum{27b}{h2_gap} points below always thinking on Qwen3.5-27B, inside the 1.5-point margin, and
inside it on the other three models. Its savings half fails: the gate escalates
\tjnum{27b}{h2_escalation} of test decisions on Qwen3.5-27B and between \tjnum{q38}{h2_escalation}
and \tjnum{9b}{h2_escalation} on the other three, against a limit of 50\% (Table~\ref{tab:scale}).
A looser tolerance does not rescue it (Table~\ref{tab:tolerance}): allowed two points below
always thinking, the gate still escalates \tjnum{27b}{gate_eps2_escalation} of decisions, and at
zero tolerance it escalates \tjnum{27b}{gate_eps0_escalation}, nearly always thinking. Confidence
gating keeps accuracy but does not let most decisions skip thinking. Figure~\ref{fig:pareto}
shows the full trade-off and Table~\ref{tab:main} breaks it down by block.

\paragraph{Rotations.} Always-S1 is the single read, without rotations. Averaged over rotations,
the read that the gate uses answers \tjnum{27b}{s1_rotation_accuracy} correctly, against
\tjnum{27b}{always_s1_accuracy} for the single read (ECE \tjnum{27b}{s1_rotation_ece} against
\tjnum{27b}{s1_raw_ece}). The single read is fragile: its most probable option changes under some
cyclic shift on \tjnum{27b}{s1_order_flip_rate} of test decisions (\tjnum{4b}{s1_order_flip_rate}
on Qwen3.5-4B). The rotations cost $K-1$ short forward passes, and they are why the gate's curve
in Figure~\ref{fig:pareto} starts above the always-S1 marker. AnyJev's batch prior, applied on top
of the rotation average, lowers accuracy to \tjnum{27b}{s1_prior_accuracy} and raises ECE to
\tjnum{27b}{s1_prior_ece}, so it is not used. Temperature scaling does most of the calibration
work: without it, the rotation average's ECE is \tjnum{27b}{s1_ece_uncalibrated}.

\begin{table}[t]
  \centering
  \small
  \begin{tabular}{@{}lrrrrr@{}}
    \toprule
    & \multicolumn{3}{c}{Gate} & \multicolumn{2}{c}{Anytime} \\
    \cmidrule(lr){2-4} \cmidrule(lr){5-6}
    Tolerance $\epsilon$ & Accuracy & Escalated & Thinking & Accuracy & Thinking \\
    \midrule
    0 points & \tjnum{27b}{gate_eps0_accuracy} & \tjnum{27b}{gate_eps0_escalation} & \tjnum{27b}{gate_eps0_share} & \tjnum{27b}{anytime_eps0_accuracy} & \tjnum{27b}{anytime_eps0_share} \\
    1 point & \tjnum{27b}{gate_eps1_accuracy} & \tjnum{27b}{gate_eps1_escalation} & \tjnum{27b}{gate_eps1_share} & \tjnum{27b}{anytime_eps1_accuracy} & \tjnum{27b}{anytime_eps1_share} \\
    2 points & \tjnum{27b}{gate_eps2_accuracy} & \tjnum{27b}{gate_eps2_escalation} & \tjnum{27b}{gate_eps2_share} & \tjnum{27b}{anytime_eps2_accuracy} & \tjnum{27b}{anytime_eps2_share} \\
    \bottomrule
  \end{tabular}
  \caption{Qwen3.5-27B, test: the gate and the anytime policy at each tolerance $\epsilon$. The
  threshold is chosen on dev to stay at most $\epsilon$ below always thinking, then applied
  unchanged to test. ``Thinking'' is thought tokens per decision as a share of always thinking's.
  The 1-point row is the setting used everywhere else. Source: \texttt{dev\_selected\_operating\_points}
  (gate), \texttt{anytime\_test.dev\_selected\_operating\_points} (anytime).}
  \label{tab:tolerance}
\end{table}

\begin{table}[t]
  \centering
  \small
  \begin{tabular}{@{}lrrrrr@{}}
    \toprule
    Block & Test decisions & Always S1 & Always S2 & Gate & Escalated \\
    \midrule
    A: routine decisions & \tjnum{27b}{block_jev_rows} & \tjnum{27b}{block_jev_s1} & \tjnum{27b}{block_jev_s2} & \tjnum{27b}{block_jev_system} & \tjnum{27b}{block_jev_escalation} \\
    B: public reasoning & \tjnum{27b}{block_public_rows} & \tjnum{27b}{block_public_s1} & \tjnum{27b}{block_public_s2} & \tjnum{27b}{block_public_system} & \tjnum{27b}{block_public_escalation} \\
    C: multi-step decisions & \tjnum{27b}{block_weak_rows} & \tjnum{27b}{block_weak_s1} & \tjnum{27b}{block_weak_s2} & \tjnum{27b}{block_weak_system} & \tjnum{27b}{block_weak_escalation} \\
    \midrule
    All & \tjnum{27b}{block_all_rows} & \tjnum{27b}{block_all_s1} & \tjnum{27b}{block_all_s2} & \tjnum{27b}{block_all_system} & \tjnum{27b}{block_all_escalation} \\
    \bottomrule
  \end{tabular}
  \caption{Qwen3.5-27B test accuracy by block, and the share of decisions the gate escalates. The
  gate's threshold is the one in \texttt{per\_family\_test\_gate} (at most one point below always
  thinking on dev, with the most decisions answered directly); values are weighted by the number of
  decisions in each entry of \texttt{per\_family\_test}. The numbers in the text use the threshold
  with the fewest thought tokens on dev
  (\texttt{anytime\_test.dev\_selected\_operating\_points}), so the ``All'' row can differ from them
  slightly. Appendix~\ref{app:full} breaks this down per source for every model.}
  \label{tab:main}
\end{table}

\subsection{Where the thinking goes}
\label{sec:where}

\begin{figure}[t]
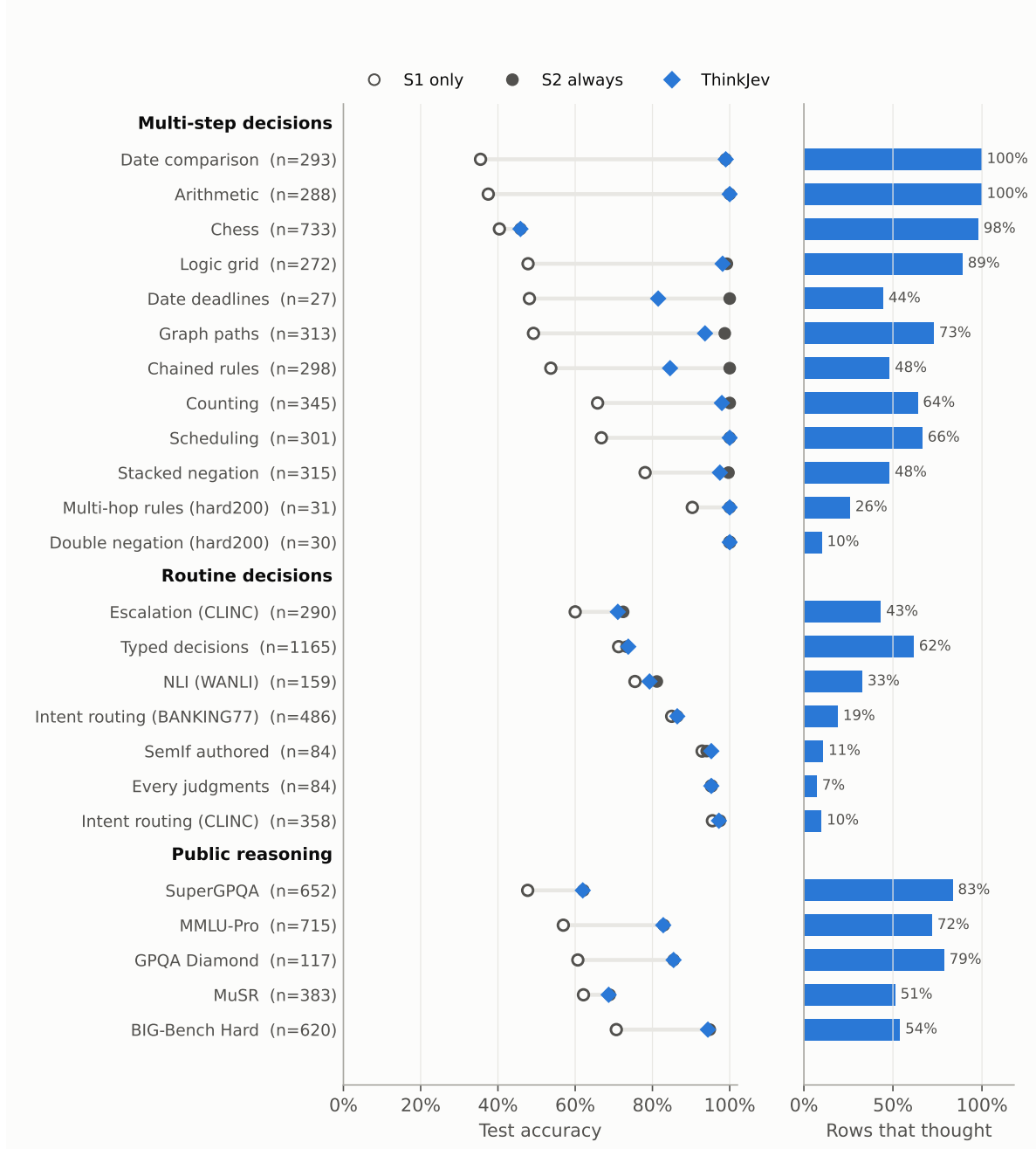

  \centering
  \tjfig{qwen3.5-27b}{tasks}{0.95\linewidth}
  \caption{Per task group: accuracy of System~1, System~2, and the gate (left), and the share of
  decisions the gate sent to System~2 (right), Qwen3.5-27B, test. Source: \texttt{per\_family\_test}.}
  \label{fig:tasks}
\end{figure}

Figure~\ref{fig:tasks} groups test decisions by task: the families of block~C and the sources of
blocks~A and~B. The gate sends most of block~C to System~2 (\tjnum{27b}{block_weak_escalation} of
its decisions), where System~1 is weakest (\tjnum{27b}{block_weak_s1} correct) and thinking helps
the most (\tjnum{27b}{block_weak_s2}). Date comparison and multi-step arithmetic are escalated on
\tjnum{27b}{grp_datecomparison_escalation} and \tjnum{27b}{grp_arithmetic_escalation} of
decisions, and System~2 answers
\tjnum{27b}{grp_datecomparison_s2} and \tjnum{27b}{grp_arithmetic_s2} of them correctly, against
\tjnum{27b}{grp_datecomparison_s1} and \tjnum{27b}{grp_arithmetic_s1} for System~1. For intent routing
on CLINC150, where System~1 is already at \tjnum{27b}{grp_intentroutingclinc_s1}, the gate escalates
only \tjnum{27b}{grp_intentroutingclinc_escalation}.

Two groups show the limit of gating on System~1's confidence. The gate escalates
\tjnum{27b}{grp_chess_escalation} of the chess puzzles, but thinking lifts their accuracy only from
\tjnum{27b}{grp_chess_s1} to \tjnum{27b}{grp_chess_s2}; on the typed-decisions dataset it escalates
\tjnum{27b}{grp_typeddecisions_escalation} for a change from \tjnum{27b}{grp_typeddecisions_s1} to
\tjnum{27b}{grp_typeddecisions_s2}. In both, System~1 is often wrong and its low confidence shows
it, yet thinking fixes few of those errors and breaks some correct answers. The gate follows where
System~1 is wrong, not where thinking helps, and that difference is what the oracle exploits
(Section~\ref{sec:h1}).

Language makes no difference within block~C. On the generated families, System~1 answers
\tjnum{27b}{lang_en_s1} of the English and \tjnum{27b}{lang_zh_s1} of the Chinese decisions, System~2
\tjnum{27b}{lang_en_s2} and \tjnum{27b}{lang_zh_s2}, and the gate escalates
\tjnum{27b}{lang_en_escalation} and \tjnum{27b}{lang_zh_escalation} (Qwen3.5-4B: System~1
\tjnum{4b}{lang_en_s1} and \tjnum{4b}{lang_zh_s1}). Chinese decisions score lower overall only because
all of them are multi-step decisions (Appendix~\ref{app:full}).

\begin{figure}[t]
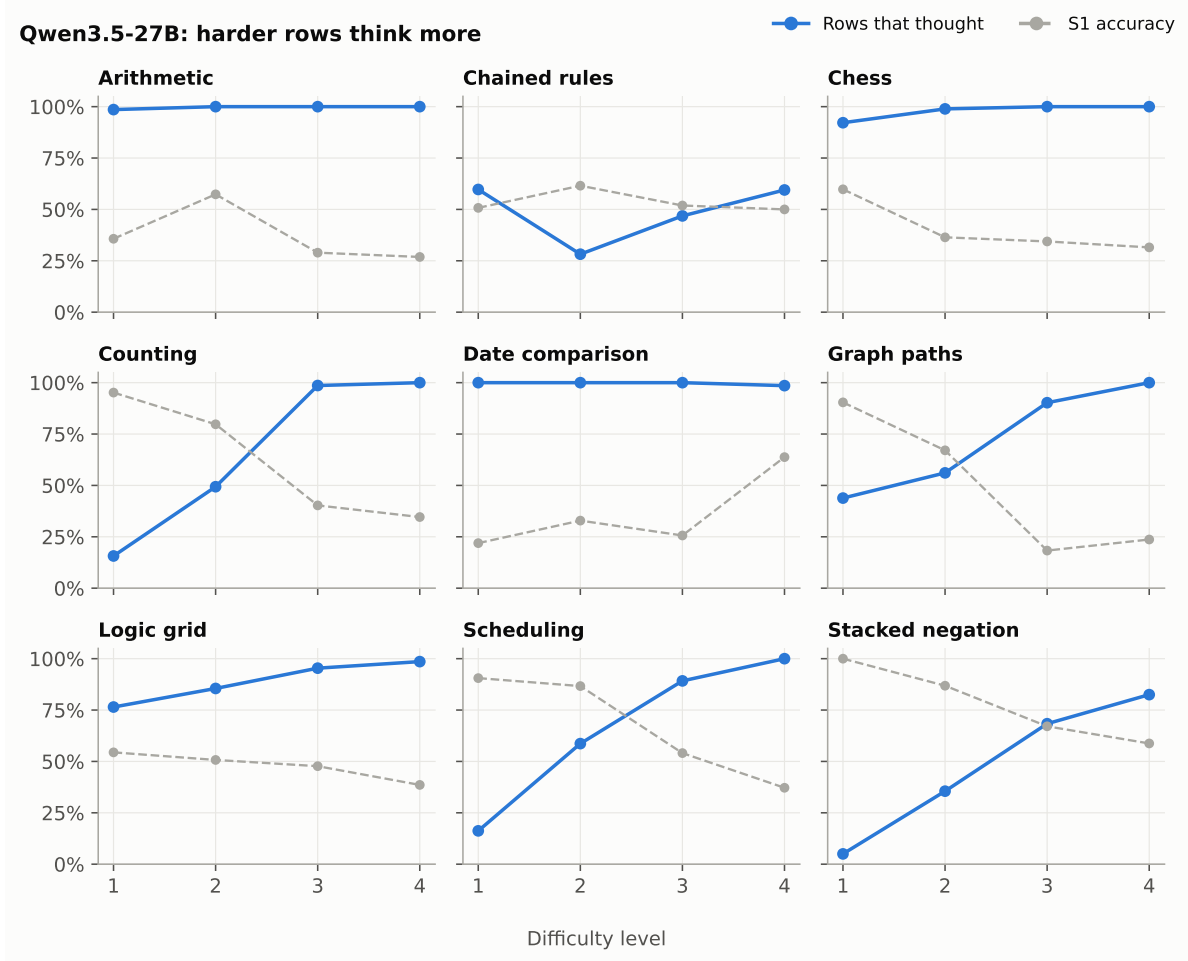

  \centering
  \tjfig{qwen3.5-27b}{difficulty}{0.95\linewidth}
  \caption{Escalation rate and System~1 accuracy by difficulty level, one panel per generated
  family, Qwen3.5-27B, test. Source: \texttt{per\_family\_test}.}
  \label{fig:difficulty}
\end{figure}

Figure~\ref{fig:difficulty} plots escalation against difficulty level for each generated family.
Escalation rises with level in \tjnum{27b}{difficulty_rising} of the \tjnum{27b}{difficulty_families}
families (rank correlation above zero), and over all generated decisions it goes from
\tjnum{27b}{weak_d1_escalation} at level~1 to \tjnum{27b}{weak_d4_escalation} at level~4, while
System~1's accuracy falls from \tjnum{27b}{weak_d1_s1} to \tjnum{27b}{weak_d4_s1}. Qwen3.5-4B shows the
same pattern in \tjnum{4b}{difficulty_rising} of \tjnum{4b}{difficulty_families} families.

\subsection{Public benchmarks}
\label{sec:public}

Eight of our sources are public benchmarks: three classification sets in block~A and five reasoning
benchmarks in block~B (Section~\ref{sec:setup}). Table~\ref{tab:public} gives System~1, System~2,
the gate, and the gate's escalation rate on each, for every model. Two patterns hold on all four
models. On the reasoning benchmarks, thinking adds the most and the gate escalates most decisions: on
Qwen3.5-27B, BIG-Bench Hard goes from \tjnum{27b}{grp_bigbenchhard_s1} with System~1 to
\tjnum{27b}{grp_bigbenchhard_s2} with System~2, MMLU-Pro from \tjnum{27b}{grp_mmlupro_s1} to
\tjnum{27b}{grp_mmlupro_s2}, and GPQA Diamond from \tjnum{27b}{grp_gpqadiamond_s1} to
\tjnum{27b}{grp_gpqadiamond_s2}, with \tjnum{27b}{grp_gpqadiamond_escalation} of GPQA Diamond
escalated. On the classification sets, thinking adds little (BANKING77:
\tjnum{27b}{grp_intentroutingbanking77_s1} to \tjnum{27b}{grp_intentroutingbanking77_s2}) and the
gate lets System~1 answer most decisions (it escalates \tjnum{27b}{grp_intentroutingclinc_escalation}
of CLINC150). Everywhere, the gate's accuracy stays within a few points of always thinking, and
closest where it escalates most, so the thinking it skips is mostly thinking that would not have
changed the answer.

These scores are not comparable with the scores reported for the same models on the same
benchmarks elsewhere. We read one option letter from SemIf's prompt (Appendix~\ref{app:prompts}),
cap thinking at 4{,}096 tokens, use one sampled thought, offer the classification sets a subset of
their labels (Appendix~\ref{app:prompts}), and score only our test split of each benchmark
(Table~\ref{tab:data}); published scores typically parse a generated answer after a longer
thought and use the full set.

\begin{table}[t]
  \centering
  \footnotesize
  \setlength{\tabcolsep}{2.4pt}
  \begingroup\renewcommand{\%}{}%
  \begin{tabular}{@{}l*{4}{rrrr}@{}}
    \toprule
    & \multicolumn{4}{c}{Qwen3.5-4B} & \multicolumn{4}{c}{Qwen3.5-9B} & \multicolumn{4}{c}{Qwen3.5-27B} & \multicolumn{4}{c}{Qwen3.8-27B} \\
    \cmidrule(lr){2-5} \cmidrule(lr){6-9} \cmidrule(lr){10-13} \cmidrule(l){14-17}
    Benchmark & S1 & S2 & Gate & Esc. & S1 & S2 & Gate & Esc. & S1 & S2 & Gate & Esc. & S1 & S2 & Gate & Esc. \\
    \midrule
    BANKING77 (intent, A) & \tjnum{4b}{grp_intentroutingbanking77_s1} & \tjnum{4b}{grp_intentroutingbanking77_s2} & \tjnum{4b}{grp_intentroutingbanking77_system} & \tjnum{4b}{grp_intentroutingbanking77_escalation} & \tjnum{9b}{grp_intentroutingbanking77_s1} & \tjnum{9b}{grp_intentroutingbanking77_s2} & \tjnum{9b}{grp_intentroutingbanking77_system} & \tjnum{9b}{grp_intentroutingbanking77_escalation} & \tjnum{27b}{grp_intentroutingbanking77_s1} & \tjnum{27b}{grp_intentroutingbanking77_s2} & \tjnum{27b}{grp_intentroutingbanking77_system} & \tjnum{27b}{grp_intentroutingbanking77_escalation} & \tjnum{q38}{grp_intentroutingbanking77_s1} & \tjnum{q38}{grp_intentroutingbanking77_s2} & \tjnum{q38}{grp_intentroutingbanking77_system} & \tjnum{q38}{grp_intentroutingbanking77_escalation} \\
    CLINC150 (intent, A) & \tjnum{4b}{grp_intentroutingclinc_s1} & \tjnum{4b}{grp_intentroutingclinc_s2} & \tjnum{4b}{grp_intentroutingclinc_system} & \tjnum{4b}{grp_intentroutingclinc_escalation} & \tjnum{9b}{grp_intentroutingclinc_s1} & \tjnum{9b}{grp_intentroutingclinc_s2} & \tjnum{9b}{grp_intentroutingclinc_system} & \tjnum{9b}{grp_intentroutingclinc_escalation} & \tjnum{27b}{grp_intentroutingclinc_s1} & \tjnum{27b}{grp_intentroutingclinc_s2} & \tjnum{27b}{grp_intentroutingclinc_system} & \tjnum{27b}{grp_intentroutingclinc_escalation} & \tjnum{q38}{grp_intentroutingclinc_s1} & \tjnum{q38}{grp_intentroutingclinc_s2} & \tjnum{q38}{grp_intentroutingclinc_system} & \tjnum{q38}{grp_intentroutingclinc_escalation} \\
    WANLI (NLI, A) & \tjnum{4b}{grp_nliwanli_s1} & \tjnum{4b}{grp_nliwanli_s2} & \tjnum{4b}{grp_nliwanli_system} & \tjnum{4b}{grp_nliwanli_escalation} & \tjnum{9b}{grp_nliwanli_s1} & \tjnum{9b}{grp_nliwanli_s2} & \tjnum{9b}{grp_nliwanli_system} & \tjnum{9b}{grp_nliwanli_escalation} & \tjnum{27b}{grp_nliwanli_s1} & \tjnum{27b}{grp_nliwanli_s2} & \tjnum{27b}{grp_nliwanli_system} & \tjnum{27b}{grp_nliwanli_escalation} & \tjnum{q38}{grp_nliwanli_s1} & \tjnum{q38}{grp_nliwanli_s2} & \tjnum{q38}{grp_nliwanli_system} & \tjnum{q38}{grp_nliwanli_escalation} \\
    \midrule
    BIG-Bench Hard (B) & \tjnum{4b}{grp_bigbenchhard_s1} & \tjnum{4b}{grp_bigbenchhard_s2} & \tjnum{4b}{grp_bigbenchhard_system} & \tjnum{4b}{grp_bigbenchhard_escalation} & \tjnum{9b}{grp_bigbenchhard_s1} & \tjnum{9b}{grp_bigbenchhard_s2} & \tjnum{9b}{grp_bigbenchhard_system} & \tjnum{9b}{grp_bigbenchhard_escalation} & \tjnum{27b}{grp_bigbenchhard_s1} & \tjnum{27b}{grp_bigbenchhard_s2} & \tjnum{27b}{grp_bigbenchhard_system} & \tjnum{27b}{grp_bigbenchhard_escalation} & \tjnum{q38}{grp_bigbenchhard_s1} & \tjnum{q38}{grp_bigbenchhard_s2} & \tjnum{q38}{grp_bigbenchhard_system} & \tjnum{q38}{grp_bigbenchhard_escalation} \\
    MMLU-Pro (B) & \tjnum{4b}{grp_mmlupro_s1} & \tjnum{4b}{grp_mmlupro_s2} & \tjnum{4b}{grp_mmlupro_system} & \tjnum{4b}{grp_mmlupro_escalation} & \tjnum{9b}{grp_mmlupro_s1} & \tjnum{9b}{grp_mmlupro_s2} & \tjnum{9b}{grp_mmlupro_system} & \tjnum{9b}{grp_mmlupro_escalation} & \tjnum{27b}{grp_mmlupro_s1} & \tjnum{27b}{grp_mmlupro_s2} & \tjnum{27b}{grp_mmlupro_system} & \tjnum{27b}{grp_mmlupro_escalation} & \tjnum{q38}{grp_mmlupro_s1} & \tjnum{q38}{grp_mmlupro_s2} & \tjnum{q38}{grp_mmlupro_system} & \tjnum{q38}{grp_mmlupro_escalation} \\
    GPQA Diamond (B) & \tjnum{4b}{grp_gpqadiamond_s1} & \tjnum{4b}{grp_gpqadiamond_s2} & \tjnum{4b}{grp_gpqadiamond_system} & \tjnum{4b}{grp_gpqadiamond_escalation} & \tjnum{9b}{grp_gpqadiamond_s1} & \tjnum{9b}{grp_gpqadiamond_s2} & \tjnum{9b}{grp_gpqadiamond_system} & \tjnum{9b}{grp_gpqadiamond_escalation} & \tjnum{27b}{grp_gpqadiamond_s1} & \tjnum{27b}{grp_gpqadiamond_s2} & \tjnum{27b}{grp_gpqadiamond_system} & \tjnum{27b}{grp_gpqadiamond_escalation} & \tjnum{q38}{grp_gpqadiamond_s1} & \tjnum{q38}{grp_gpqadiamond_s2} & \tjnum{q38}{grp_gpqadiamond_system} & \tjnum{q38}{grp_gpqadiamond_escalation} \\
    SuperGPQA (B) & \tjnum{4b}{grp_supergpqa_s1} & \tjnum{4b}{grp_supergpqa_s2} & \tjnum{4b}{grp_supergpqa_system} & \tjnum{4b}{grp_supergpqa_escalation} & \tjnum{9b}{grp_supergpqa_s1} & \tjnum{9b}{grp_supergpqa_s2} & \tjnum{9b}{grp_supergpqa_system} & \tjnum{9b}{grp_supergpqa_escalation} & \tjnum{27b}{grp_supergpqa_s1} & \tjnum{27b}{grp_supergpqa_s2} & \tjnum{27b}{grp_supergpqa_system} & \tjnum{27b}{grp_supergpqa_escalation} & \tjnum{q38}{grp_supergpqa_s1} & \tjnum{q38}{grp_supergpqa_s2} & \tjnum{q38}{grp_supergpqa_system} & \tjnum{q38}{grp_supergpqa_escalation} \\
    MuSR (B) & \tjnum{4b}{grp_musr_s1} & \tjnum{4b}{grp_musr_s2} & \tjnum{4b}{grp_musr_system} & \tjnum{4b}{grp_musr_escalation} & \tjnum{9b}{grp_musr_s1} & \tjnum{9b}{grp_musr_s2} & \tjnum{9b}{grp_musr_system} & \tjnum{9b}{grp_musr_escalation} & \tjnum{27b}{grp_musr_s1} & \tjnum{27b}{grp_musr_s2} & \tjnum{27b}{grp_musr_system} & \tjnum{27b}{grp_musr_escalation} & \tjnum{q38}{grp_musr_s1} & \tjnum{q38}{grp_musr_s2} & \tjnum{q38}{grp_musr_system} & \tjnum{q38}{grp_musr_escalation} \\
    \bottomrule
  \end{tabular}
  \endgroup
  \caption{Public benchmarks, test split: accuracy (\%) of System~1, System~2, and the gate, and the
  share of decisions the gate escalates (\%), for every model. The gate uses the threshold in
  \texttt{per\_family\_test\_gate}, as in Table~\ref{tab:main}; scores follow our
  protocol (one option letter, a 4{,}096-token budget) and are not comparable with published
  scores. Source: \texttt{per\_family\_test}.}
  \label{tab:public}
\end{table}

\subsection{The anytime curve}
\label{sec:anytime}

\begin{figure}[t]
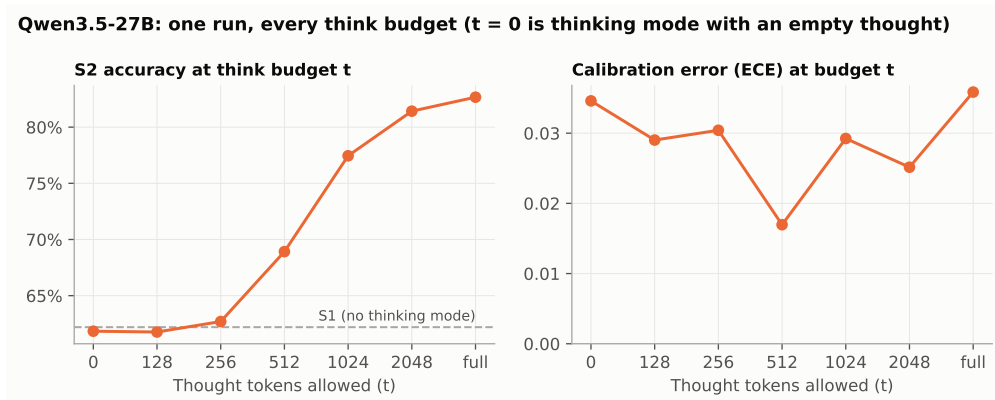

  \centering
  \tjfig{qwen3.5-27b}{anytime}{0.8\linewidth}
  \caption{System~2 accuracy and expected calibration error under each thinking budget $t$, read from
  one sampled thought per decision, Qwen3.5-27B, test. Source: \texttt{anytime\_test.per\_checkpoint},
  \texttt{anytime\_test.full\_thought}.}
  \label{fig:anytime}
\end{figure}

Figure~\ref{fig:anytime} shows test accuracy and calibration error of System~2 at each think
budget, each with its own dev temperature. At $t = 0$, the thinking template with an empty thought
reaches \tjnum{27b}{anytime_acc_t0} (ECE \tjnum{27b}{anytime_ece_t0}), next to
\tjnum{27b}{s1_raw_accuracy} (ECE \tjnum{27b}{s1_raw_ece}) for System~1 read without rotations: the
two reads nearly coincide, as Section~\ref{sec:anytime-method} expects. The curve stays flat over
the first 256 thought tokens (\tjnum{27b}{anytime_acc_t256}), rises through 512
(\tjnum{27b}{anytime_acc_t512}) and 1{,}024 (\tjnum{27b}{anytime_acc_t1024}) tokens, and is at
\tjnum{27b}{anytime_acc_t2048} by 2{,}048, close to the full thought's
\tjnum{27b}{anytime_acc_full}; \tjnum{27b}{s2_closed_naturally} of thoughts end by themselves within
the 4{,}096-token budget. Mid-thought reads stay calibrated: ECE lies between
\tjnum{27b}{anytime_ece_t512} and \tjnum{27b}{anytime_ece_full} at every checkpoint. They stay
calibrated only because each checkpoint has its own temperature. The fitted temperature rises
with the budget on every model, from \tjnum{27b}{temp_t0} at $t = 0$ to \tjnum{27b}{temp_full} at the
full thought on Qwen3.5-27B (Qwen3.5-4B: \tjnum{4b}{temp_t0} to \tjnum{4b}{temp_full};
Qwen3.8-27B: \tjnum{q38}{temp_t0} to \tjnum{q38}{temp_full}), against \tjnum{27b}{temp_s1} for
System~1's rotation average. The raw logits grow more confident faster than the answers improve: the longer
the model has thought, the more its raw confidence overstates its accuracy, and one temperature
cannot serve every budget. The Qwen3.5 models share the shape of the accuracy curve. Qwen3.8-27B thinks far less (\tjnum{q38}{always_s2_mean_think_tokens} mean
thought tokens), so its curve rises earlier (\tjnum{q38}{anytime_acc_t256} at 256 tokens), and its
calibration error is higher at long budgets, from \tjnum{q38}{anytime_ece_t0} at $t = 0$ to
\tjnum{q38}{anytime_ece_full} at the full thought. We report the anytime curve descriptively and
draw no directional conclusion from it.

Reading every budget from one thought is what makes the anytime policy possible: on
Qwen3.5-27B it spends \tjnum{27b}{anytime_thinking_share} of always thinking's thought tokens,
against the gate's \tjnum{27b}{gate_thinking_share}, at the same accuracy. The saving comes from
stopping early. With $\tau$ chosen on dev, the policy answers \tjnum{27b}{stop_s1} of test
decisions with System~1, \tjnum{27b}{stop_checkpoint} at a checkpoint before the full thought, and
\tjnum{27b}{stop_full} after the full thought; \tjnum{27b}{stop_checkpoint_of_escalated} of the
escalated decisions stop early. The saving shrinks where thoughts are short: on Qwen3.8-27B only
\tjnum{q38}{stop_checkpoint_of_escalated} of escalated decisions stop early, because few checkpoints
fall before the end of its thoughts, and the two policies spend
\tjnum{q38}{anytime_thinking_share} and \tjnum{q38}{gate_thinking_share}.

\subsection{Risk guarantee}
\label{sec:risk}

\begin{figure}[t]
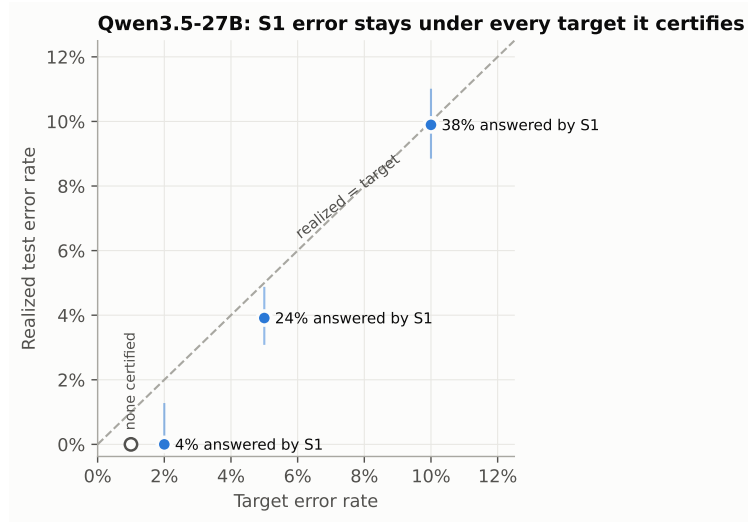

  \centering
  \tjfig{qwen3.5-27b}{risk}{0.6\linewidth}
  \caption{Target risk against realized test risk of System~1 answers accepted under the
  Learn-then-Test threshold ($\delta = 0.05$), Qwen3.5-27B, with exact binomial 95\% intervals.
  The hollow marker is a target at which no decision is accepted. Source:
  \texttt{s1\_readouts\_test.rotation.coverage\_at\_risk}.}
  \label{fig:risk}
\end{figure}

At a 5\% target risk, the Learn-then-Test threshold chosen on dev lets System~1 answer
\tjnum{27b}{s1_ltt_coverage_at_5pct} of Qwen3.5-27B's test decisions, and those answers have a
test error rate (realized risk) of \tjnum{27b}{s1_ltt_test_risk_at_5pct} \tjnum{27b}{s1_ltt_interval_at_5pct}.
Table~\ref{tab:risk} gives all four targets for every model and Figure~\ref{fig:risk} plots them for
Qwen3.5-27B.

On the three Qwen3.5 models, realized risk stays at or under the target wherever the threshold
accepts any decision, so the risk criterion holds. It holds only trivially where the threshold
accepts nothing, because then System~1 answers nothing. Qwen3.5-27B accepts decisions at
\tjnum{27b}{s1_ltt_targets_certified} of the four targets (all but 1\%), Qwen3.5-4B at
\tjnum{4b}{s1_ltt_targets_certified} (5\% and 10\%), and Qwen3.5-9B at
\tjnum{9b}{s1_ltt_targets_certified} (10\% only). Accepting nothing follows from the test design:
fixed-sequence testing starts at the strictest threshold and stops at its first failure, and with
groups as the sample size a 1\% target needs about 300 accepted groups before even an error-free
set of accepted decisions can pass. If System~1's most confident dev answers already contain errors, the sequence stops
before accepting anything. On Qwen3.8-27B the threshold accepts decisions at all
four targets, but at the 10\% target the realized risk is \tjnum{q38}{s1_ltt_test_risk_at_10pct}
\tjnum{q38}{s1_ltt_interval_at_10pct}, above the target. Its interval contains the target, and
Learn-then-Test bounds risk only with probability $1 - \delta$ over dev draws, but by the risk
criterion this is a violation, and we report it as one.

\begin{table}[t]
  \centering
  \small
  \begin{tabular}{@{}llllll@{}}
    \toprule
    Target & & Qwen3.5-4B & Qwen3.5-9B & Qwen3.5-27B & Qwen3.8-27B \\
    \midrule
    1\% & answered by S1 & \tjnum{4b}{s1_ltt_coverage_at_1pct} & \tjnum{9b}{s1_ltt_coverage_at_1pct} & \tjnum{27b}{s1_ltt_coverage_at_1pct} & \tjnum{q38}{s1_ltt_coverage_at_1pct} \\
        & realized risk & \tjnum{4b}{s1_ltt_test_risk_at_1pct} \tjnum{4b}{s1_ltt_interval_at_1pct} & \tjnum{9b}{s1_ltt_test_risk_at_1pct} \tjnum{9b}{s1_ltt_interval_at_1pct} & \tjnum{27b}{s1_ltt_test_risk_at_1pct} \tjnum{27b}{s1_ltt_interval_at_1pct} & \tjnum{q38}{s1_ltt_test_risk_at_1pct} \tjnum{q38}{s1_ltt_interval_at_1pct} \\
    2\% & answered by S1 & \tjnum{4b}{s1_ltt_coverage_at_2pct} & \tjnum{9b}{s1_ltt_coverage_at_2pct} & \tjnum{27b}{s1_ltt_coverage_at_2pct} & \tjnum{q38}{s1_ltt_coverage_at_2pct} \\
        & realized risk & \tjnum{4b}{s1_ltt_test_risk_at_2pct} \tjnum{4b}{s1_ltt_interval_at_2pct} & \tjnum{9b}{s1_ltt_test_risk_at_2pct} \tjnum{9b}{s1_ltt_interval_at_2pct} & \tjnum{27b}{s1_ltt_test_risk_at_2pct} \tjnum{27b}{s1_ltt_interval_at_2pct} & \tjnum{q38}{s1_ltt_test_risk_at_2pct} \tjnum{q38}{s1_ltt_interval_at_2pct} \\
    5\% & answered by S1 & \tjnum{4b}{s1_ltt_coverage_at_5pct} & \tjnum{9b}{s1_ltt_coverage_at_5pct} & \tjnum{27b}{s1_ltt_coverage_at_5pct} & \tjnum{q38}{s1_ltt_coverage_at_5pct} \\
        & realized risk & \tjnum{4b}{s1_ltt_test_risk_at_5pct} \tjnum{4b}{s1_ltt_interval_at_5pct} & \tjnum{9b}{s1_ltt_test_risk_at_5pct} \tjnum{9b}{s1_ltt_interval_at_5pct} & \tjnum{27b}{s1_ltt_test_risk_at_5pct} \tjnum{27b}{s1_ltt_interval_at_5pct} & \tjnum{q38}{s1_ltt_test_risk_at_5pct} \tjnum{q38}{s1_ltt_interval_at_5pct} \\
    10\% & answered by S1 & \tjnum{4b}{s1_ltt_coverage_at_10pct} & \tjnum{9b}{s1_ltt_coverage_at_10pct} & \tjnum{27b}{s1_ltt_coverage_at_10pct} & \tjnum{q38}{s1_ltt_coverage_at_10pct} \\
        & realized risk & \tjnum{4b}{s1_ltt_test_risk_at_10pct} \tjnum{4b}{s1_ltt_interval_at_10pct} & \tjnum{9b}{s1_ltt_test_risk_at_10pct} \tjnum{9b}{s1_ltt_interval_at_10pct} & \tjnum{27b}{s1_ltt_test_risk_at_10pct} \tjnum{27b}{s1_ltt_interval_at_10pct} & \tjnum{q38}{s1_ltt_test_risk_at_10pct} \tjnum{q38}{s1_ltt_interval_at_10pct} \\
    \bottomrule
  \end{tabular}
  \caption{System~1 answers accepted by the Learn-then-Test threshold ($\delta = 0.05$), test: the
  share of decisions answered by System~1 and the error rate of those answers (realized risk), with
  an exact binomial 95\% interval over the accepted decisions. ``0.0\%'' answered with ``--'' risk
  means the threshold accepts no decision. Source:
  \texttt{s1\_readouts\_test.rotation.coverage\_at\_risk}.}
  \label{tab:risk}
\end{table}

\subsection{Scale}
\label{sec:scale}

Table~\ref{tab:scale} repeats the headline quantities for all four models with the same settings.
Across the Qwen3.5 models, accuracy rises with scale for both systems, from
\tjnum{4b}{always_s1_accuracy} to \tjnum{27b}{always_s1_accuracy} for System~1 and from
\tjnum{4b}{always_s2_accuracy} to \tjnum{27b}{always_s2_accuracy} for System~2, and the gap
that thinking closes narrows. The risk guarantee does not improve steadily with scale: Qwen3.5-9B
accepts decisions at fewer targets than Qwen3.5-4B. Qwen3.8-27B is close to Qwen3.5-27B in accuracy with less than half
the thinking, which leaves the anytime policy little to cut.

\begin{table}[t]
  \centering
  \small
  \begin{tabular}{@{}lrrrr@{}}
    \toprule
    Quantity & Qwen3.5-4B & Qwen3.5-9B & Qwen3.5-27B & Qwen3.8-27B \\
    \midrule
    Test decisions & \tjnum{4b}{rows_test} & \tjnum{9b}{rows_test} & \tjnum{27b}{rows_test} & \tjnum{q38}{rows_test} \\
    Always S1: accuracy & \tjnum{4b}{always_s1_accuracy} & \tjnum{9b}{always_s1_accuracy} & \tjnum{27b}{always_s1_accuracy} & \tjnum{q38}{always_s1_accuracy} \\
    Always S2: accuracy & \tjnum{4b}{always_s2_accuracy} & \tjnum{9b}{always_s2_accuracy} & \tjnum{27b}{always_s2_accuracy} & \tjnum{q38}{always_s2_accuracy} \\
    Always S2: thought tokens per decision & \tjnum{4b}{always_s2_mean_think_tokens} & \tjnum{9b}{always_s2_mean_think_tokens} & \tjnum{27b}{always_s2_mean_think_tokens} & \tjnum{q38}{always_s2_mean_think_tokens} \\
    Oracle accuracy & \tjnum{4b}{oracle_accuracy} & \tjnum{9b}{oracle_accuracy} & \tjnum{27b}{oracle_accuracy} & \tjnum{q38}{oracle_accuracy} \\
    \midrule
    Gate: accuracy & \tjnum{4b}{gate_test_accuracy} & \tjnum{9b}{gate_test_accuracy} & \tjnum{27b}{gate_test_accuracy} & \tjnum{q38}{gate_test_accuracy} \\
    Gate: share of always-S2 tokens & \tjnum{4b}{gate_thinking_share} & \tjnum{9b}{gate_thinking_share} & \tjnum{27b}{gate_thinking_share} & \tjnum{q38}{gate_thinking_share} \\
    Savings: points below always S2 (limit 1.5) & \tjnum{4b}{h2_gap} & \tjnum{9b}{h2_gap} & \tjnum{27b}{h2_gap} & \tjnum{q38}{h2_gap} \\
    Savings: escalated (limit 50\%) & \tjnum{4b}{h2_escalation} & \tjnum{9b}{h2_escalation} & \tjnum{27b}{h2_escalation} & \tjnum{q38}{h2_escalation} \\
    Anytime: accuracy & \tjnum{4b}{anytime_test_accuracy} & \tjnum{9b}{anytime_test_accuracy} & \tjnum{27b}{anytime_test_accuracy} & \tjnum{q38}{anytime_test_accuracy} \\
    Anytime: share of always-S2 tokens & \tjnum{4b}{anytime_thinking_share} & \tjnum{9b}{anytime_thinking_share} & \tjnum{27b}{anytime_thinking_share} & \tjnum{q38}{anytime_thinking_share} \\
    \midrule
    System~1 ECE (calibrated) & \tjnum{4b}{s1_ece} & \tjnum{9b}{s1_ece} & \tjnum{27b}{s1_ece} & \tjnum{q38}{s1_ece} \\
    LTT: targets with any acceptance (of 4) & \tjnum{4b}{s1_ltt_targets_certified} & \tjnum{9b}{s1_ltt_targets_certified} & \tjnum{27b}{s1_ltt_targets_certified} & \tjnum{q38}{s1_ltt_targets_certified} \\
    LTT at 5\%: answered by S1 & \tjnum{4b}{s1_ltt_coverage_at_5pct} & \tjnum{9b}{s1_ltt_coverage_at_5pct} & \tjnum{27b}{s1_ltt_coverage_at_5pct} & \tjnum{q38}{s1_ltt_coverage_at_5pct} \\
    LTT at 5\%: realized risk & \tjnum{4b}{s1_ltt_test_risk_at_5pct} & \tjnum{9b}{s1_ltt_test_risk_at_5pct} & \tjnum{27b}{s1_ltt_test_risk_at_5pct} & \tjnum{q38}{s1_ltt_test_risk_at_5pct} \\
    \bottomrule
  \end{tabular}
  \caption{Headline quantities for every model, test. Always S1: always answer directly; always
  S2: always think. Gate and anytime values use the thresholds chosen on dev
  (\texttt{dev\_selected\_operating\_points}, tolerance one point). ``Savings'' rows are the two
  halves of the savings criterion; LTT rows are the Learn-then-Test threshold for System~1. Each value's report
  field is named in \texttt{results/\allowbreak figures/\allowbreak <model>/\allowbreak numbers.tex}.}
  \label{tab:scale}
\end{table}

\subsection{Gating versus pre-routing}
\label{sec:h1}

A good policy escalates first the decisions where thinking helps. Table~\ref{tab:h1} measures this
with the normalized area under the accuracy--coverage curve (Section~\ref{sec:setup}), comparing the
gate with each pre-router. A positive difference favors the gate, and we claim a difference only
when its bootstrap interval excludes zero. The zero-shot router, which asks the model whether a
decision can be made at a glance, ranks decisions better than the gate on Qwen3.5-9B, Qwen3.5-27B,
and Qwen3.8-27B, and ties with it on Qwen3.5-4B. The probe router, which reads the model's hidden
states, is worse than the gate on Qwen3.5-4B and Qwen3.5-27B and ties on the other two models. The $n$-gram
router is worse everywhere. All policies stay far from the oracle: the best normalized area is
\tjnum{27b}{area_router_zero_shot} on Qwen3.5-27B, so no policy captures even half of the oracle's
advantage over random escalation. Section~\ref{sec:where} suggests why a System~1 signal cannot:
the decisions where System~1 is least sure include many that thinking does not fix. The zero-shot
router costs one more short read on the same prefix, so it is as cheap as the gate. The probe
router appears in Figure~\ref{fig:pareto} for reference.

\paragraph{Combining the two signals (post hoc).} After these results had been read, we fit a
logistic regression on the logits of the gate's confidence and the zero-shot score, trained on dev
to predict System~1's correctness and cross-fitted by group (Appendix~\ref{app:repro}). It is
ahead of the gate on every model by normalized area (Qwen3.5-27B: \tjnum{27b}{area_combined}
against \tjnum{27b}{area_confidence_rot}, difference \tjnum{27b}{area_diff_combined}
\tjnum{27b}{area_ci_combined}; last row of Table~\ref{tab:h1}), but on the three larger models it
stays below the zero-shot router alone (\tjnum{27b}{area_router_zero_shot} on Qwen3.5-27B). The
fit says why. Trained to predict whether System~1 is right, it puts almost all of its weight on
the confidence (standardized weights \tjnum{27b}{combined_weight_confidence_rot} and
\tjnum{27b}{combined_weight_router_zero_shot} on Qwen3.5-27B) and matches the confidence's AUROC
for System~1 correctness (\tjnum{27b}{auroc_combined} against \tjnum{27b}{auroc_confidence_rot};
the zero-shot score alone has \tjnum{27b}{auroc_router_zero_shot}). The zero-shot router ranks
worse by that target and better by area because its low scores pick out the decisions that
thinking fixes, not the decisions System~1 gets wrong. With its threshold chosen on dev, the
combination escalates \tjnum{27b}{combined_eps1_escalation} of decisions against the gate's
\tjnum{27b}{h2_escalation} (Qwen3.8-27B: \tjnum{q38}{combined_eps1_escalation} against
\tjnum{q38}{h2_escalation}), so it does not change the verdict on the savings criterion. A combination trained on the
gain from thinking needs System~2 labels on dev and is left to future work.

\begin{table}[t]
  \centering
  \small
  \begin{tabular}{@{}lllll@{}}
    \toprule
    & Qwen3.5-4B & Qwen3.5-9B & Qwen3.5-27B & Qwen3.8-27B \\
    \midrule
    Gate: normalized area & \tjnum{4b}{area_confidence_rot} & \tjnum{9b}{area_confidence_rot} & \tjnum{27b}{area_confidence_rot} & \tjnum{q38}{area_confidence_rot} \\
    \midrule
    Gate $-$ zero-shot router & \tjnum{4b}{area_diff_router_zero_shot} & \tjnum{9b}{area_diff_router_zero_shot} & \tjnum{27b}{area_diff_router_zero_shot} & \tjnum{q38}{area_diff_router_zero_shot} \\
    & \tjnum{4b}{area_ci_router_zero_shot} & \tjnum{9b}{area_ci_router_zero_shot} & \tjnum{27b}{area_ci_router_zero_shot} & \tjnum{q38}{area_ci_router_zero_shot} \\
    Gate $-$ probe router & \tjnum{4b}{area_diff_router_probe} & \tjnum{9b}{area_diff_router_probe} & \tjnum{27b}{area_diff_router_probe} & \tjnum{q38}{area_diff_router_probe} \\
    & \tjnum{4b}{area_ci_router_probe} & \tjnum{9b}{area_ci_router_probe} & \tjnum{27b}{area_ci_router_probe} & \tjnum{q38}{area_ci_router_probe} \\
    Gate $-$ $n$-gram router & \tjnum{4b}{area_diff_router_supervised} & \tjnum{9b}{area_diff_router_supervised} & \tjnum{27b}{area_diff_router_supervised} & \tjnum{q38}{area_diff_router_supervised} \\
    & \tjnum{4b}{area_ci_router_supervised} & \tjnum{9b}{area_ci_router_supervised} & \tjnum{27b}{area_ci_router_supervised} & \tjnum{q38}{area_ci_router_supervised} \\
    \midrule
    Gate $-$ combination & \tjnum{4b}{area_diff_combined} & \tjnum{9b}{area_diff_combined} & \tjnum{27b}{area_diff_combined} & \tjnum{q38}{area_diff_combined} \\
    & \tjnum{4b}{area_ci_combined} & \tjnum{9b}{area_ci_combined} & \tjnum{27b}{area_ci_combined} & \tjnum{q38}{area_ci_combined} \\
    \bottomrule
  \end{tabular}
  \caption{Ranking comparison: normalized area under the test accuracy--coverage curve,
  $(\text{policy} - \text{random}) / (\text{oracle} - \text{random})$, of the gate, and the gate's
  difference from each pre-router with a 95\% interval from a bootstrap over groups (1{,}000
  resamples); a negative difference means the router ranks better. The
  last row is the combination of the gate's confidence and the zero-shot score, fit post hoc
  (Section~\ref{sec:h1}). Source: \texttt{normalized\_area\_test}.}
  \label{tab:h1}
\end{table}

\subsection{Latency}
\label{sec:latency}

On one H200 at batch size 1, with a median shared prefix of \tjnum{4b}{lat_prefix_tokens}
tokens, Qwen3.5-4B spends \tjnum{4b}{lat_prefill_ms}~ms processing the shared prefix (the
prefill) and \tjnum{4b}{lat_s1_marginal_ms}~ms copying the cache and reading System~1's answer
slots; the rotations behind the gate's confidence add \tjnum{4b}{lat_rotation_ms}~ms. An escalated
decision continues from the same prefill, so sharing the prefix saves one prefill per escalation.
That saving is small next to thinking itself: System~2 takes a median of
\tjnum{4b}{lat_s2_seconds}~s (\tjnum{4b}{lat_s2_think_tokens} thought tokens) on the same
decisions. Whether a decision thinks, not the prefill, decides how long it takes. Latency was
measured on Qwen3.5-4B only, over \tjnum{4b}{lat_rows} stratified test decisions, all of them
thinking.

\subsection{External check on JevBench}
\label{sec:jevbench}

Our own test set cannot compare ThinkJev with commercial decision services, because we do not call
their APIs. JevBench~\citep{jevbench2026}, an open benchmark of typed-decision services maintained
independently of us, makes that comparison possible: it publishes, next to its items, each listed
system's outcome on every item. Its \tjnum{27b}{jb_rows} public decisions come in three tiers. The
easy and standard tiers hold short business decisions (intent and ticket routing, fact checks,
field extraction, tool selection, policy compliance, adequacy and ordinal ratings); the
\tjnum{27b}{jb_hard_rows} hard items add long policies, multi-hop and temporal or numeric
reasoning, probability estimates, ambiguous trade-offs, and adversarial phrasing. Most are choice
questions; the rest are yes/no and score questions. The listed systems include Jev~1.13.0, the
thinking APIs DeepSeek V4.1 Flash and GPT-5.6 Luna, and four further systems, among them SemIf on
Qwen3.5-4B (Table~\ref{tab:jevbench}).

Each item is sent as the request JevBench's own harness builds (a state, instructions, and labeled
options), turned into a decision by the same mapping the server applies
(Appendix~\ref{app:server}), and read with the same prompts as every other decision (Appendix~\ref{app:prompts} shows one). It is
scored by JevBench's rule: the top label must equal the expected one. Nothing is fit on JevBench:
the temperatures and the gate and anytime thresholds are each model's values chosen on our dev
split (Section~\ref{sec:main}), so the check also tests whether they transfer to a new domain. This check
was added after the main results had been read and is exploratory. The easy and standard items are
near ceiling for most listed systems, so differences come from the hard tier.
Table~\ref{tab:jevbench} gives the accuracies, and we compare each policy with each listed system by
an exact McNemar test on the same items.

System~1 alone is level with Jev. Qwen3.5-27B's System~1 answers \tjnum{27b}{jb_s1_correct} of
\tjnum{27b}{jb_rows} items correctly against Jev's \tjnum{27b}{jb_sys_jev1130_correct}
(\tjnum{27b}{jb_s1_vs_jev1130_all_ours} items only ThinkJev gets right,
\tjnum{27b}{jb_s1_vs_jev1130_all_theirs} only Jev, \tjnum{27b}{jb_s1_vs_jev1130_all_p}); Qwen3.5-4B's
System~1 is below Jev (\tjnum{4b}{jb_s1_correct}, \tjnum{4b}{jb_s1_vs_jev1130_all_p}), close to
SemIf's own entry on the same model (\tjnum{27b}{jb_sys_semifqwen354b_correct}), as expected from the
same prompt. Thinking where System~1 is unsure changes this. On Qwen3.5-27B the gate answers
\tjnum{27b}{jb_gate_correct} items (\tjnum{27b}{jb_gate_vs_jev1130_all_ours} only ThinkJev,
\tjnum{27b}{jb_gate_vs_jev1130_all_theirs} only Jev, \tjnum{27b}{jb_gate_vs_jev1130_all_p}) while
thinking on \tjnum{27b}{jb_gate_escalation} of them, \tjnum{27b}{jb_gate_think} thought tokens per
item against \tjnum{27b}{jb_s2_think} for always thinking (\tjnum{27b}{jb_s2_correct} correct). Every
thinking policy is above Jev on every model, on all items and on the hard tier alone, with the
smallest margin on Qwen3.5-4B (gate \tjnum{4b}{jb_gate_correct}, \tjnum{4b}{jb_gate_vs_jev1130_all_p}).

The most accurate listed systems are two thinking APIs, DeepSeek V4.1 Flash with its default
thinking (\tjnum{27b}{jb_sys_deepseekflash_correct}) and GPT-5.6 Luna at low reasoning effort
(\tjnum{27b}{jb_sys_gpt56luna_correct}). Qwen3.5-27B's gate cannot be told apart from either
(\tjnum{27b}{jb_gate_vs_deepseekflash_all_p} and \tjnum{27b}{jb_gate_vs_gpt56luna_all_p}), nor can
Qwen3.8-27B's (\tjnum{q38}{jb_gate_correct}; \tjnum{q38}{jb_gate_vs_deepseekflash_all_p} and
\tjnum{q38}{jb_gate_vs_gpt56luna_all_p}). Qwen3.5-4B's gate is below both
(\tjnum{4b}{jb_gate_vs_deepseekflash_all_p} and \tjnum{4b}{jb_gate_vs_gpt56luna_all_p});
Qwen3.5-9B's is not distinguishable from them over all items
(\tjnum{9b}{jb_gate_vs_deepseekflash_all_p} and \tjnum{9b}{jb_gate_vs_gpt56luna_all_p}) but below
them on the hard tier. So an open 27B model, without training and thinking on fewer than a third
of the decisions, is in the same accuracy band as these APIs here. ``Cannot be told apart'' is not
a tie: with near-ceiling accuracy and \tjnum{27b}{jb_hard_rows} hard items, the test resolves only
large gaps.

Unlike on our test set, the gate thinks on a minority of these decisions, between
\tjnum{q38}{jb_gate_escalation} and \tjnum{9b}{jb_gate_escalation} across the four models, presumably
because JevBench holds fewer multi-step decisions than our block~C. Three caveats. The listed
systems were measured by JevBench, through their own servers and at their own settings, not by us.
JevBench's own score also weighs speed and cost, and we report accuracy only: an escalated
decision takes seconds (Section~\ref{sec:latency}) where Jev answers in under one. Contamination is
unlikely: JevBench froze its items on 2026-09-19, after every model version we use was released (2026-08-14 for Qwen3.8-27B, March and April 2026 for the Qwen3.5 models).

\begin{table}[t]
  \centering
  \small
  \begin{tabular}{@{}lllll@{}}
    \toprule
    Policy & Qwen3.5-4B & Qwen3.5-9B & Qwen3.5-27B & Qwen3.8-27B \\
    \midrule
    S1 only & \tjnum{4b}{jb_s1_acc} / \tjnum{4b}{jb_s1_hard_acc} & \tjnum{9b}{jb_s1_acc} / \tjnum{9b}{jb_s1_hard_acc} & \tjnum{27b}{jb_s1_acc} / \tjnum{27b}{jb_s1_hard_acc} & \tjnum{q38}{jb_s1_acc} / \tjnum{q38}{jb_s1_hard_acc} \\
    Gate & \tjnum{4b}{jb_gate_acc} / \tjnum{4b}{jb_gate_hard_acc} & \tjnum{9b}{jb_gate_acc} / \tjnum{9b}{jb_gate_hard_acc} & \tjnum{27b}{jb_gate_acc} / \tjnum{27b}{jb_gate_hard_acc} & \tjnum{q38}{jb_gate_acc} / \tjnum{q38}{jb_gate_hard_acc} \\
    Anytime & \tjnum{4b}{jb_anytime_acc} / \tjnum{4b}{jb_anytime_hard_acc} & \tjnum{9b}{jb_anytime_acc} / \tjnum{9b}{jb_anytime_hard_acc} & \tjnum{27b}{jb_anytime_acc} / \tjnum{27b}{jb_anytime_hard_acc} & \tjnum{q38}{jb_anytime_acc} / \tjnum{q38}{jb_anytime_hard_acc} \\
    Always S2 & \tjnum{4b}{jb_s2_acc} / \tjnum{4b}{jb_s2_hard_acc} & \tjnum{9b}{jb_s2_acc} / \tjnum{9b}{jb_s2_hard_acc} & \tjnum{27b}{jb_s2_acc} / \tjnum{27b}{jb_s2_hard_acc} & \tjnum{q38}{jb_s2_acc} / \tjnum{q38}{jb_s2_hard_acc} \\
    Escalated by the gate & \tjnum{4b}{jb_gate_escalation} & \tjnum{9b}{jb_gate_escalation} & \tjnum{27b}{jb_gate_escalation} & \tjnum{q38}{jb_gate_escalation} \\
    \midrule
    \multicolumn{5}{@{}l}{Listed on JevBench (same items, from its per-task file)} \\
    \quad Jev 1.13.0 & \multicolumn{4}{l}{\tjnum{27b}{jb_sys_jev1130_acc} / \tjnum{27b}{jb_sys_jev1130_hard_acc}} \\
    \quad DeepSeek V4.1 Flash (thinking) & \multicolumn{4}{l}{\tjnum{27b}{jb_sys_deepseekflash_acc} / \tjnum{27b}{jb_sys_deepseekflash_hard_acc}} \\
    \quad GPT-5.6 Luna (low reasoning) & \multicolumn{4}{l}{\tjnum{27b}{jb_sys_gpt56luna_acc} / \tjnum{27b}{jb_sys_gpt56luna_hard_acc}} \\
    \quad OpenJev (thinking) & \multicolumn{4}{l}{\tjnum{27b}{jb_sys_openjevthinking_acc} / \tjnum{27b}{jb_sys_openjevthinking_hard_acc}} \\
    \quad djev (thinking) & \multicolumn{4}{l}{\tjnum{27b}{jb_sys_djevthinking_acc} / \tjnum{27b}{jb_sys_djevthinking_hard_acc}} \\
    \quad SemIf (Qwen3.5-4B) & \multicolumn{4}{l}{\tjnum{27b}{jb_sys_semifqwen354b_acc} / \tjnum{27b}{jb_sys_semifqwen354b_hard_acc}} \\
    \quad reflex-27b (Qwen3.8-27B) & \multicolumn{4}{l}{\tjnum{27b}{jb_sys_reflex27b_acc} / \tjnum{27b}{jb_sys_reflex27b_hard_acc}} \\
    \bottomrule
  \end{tabular}
  \caption{JevBench public items: accuracy on all \tjnum{27b}{jb_rows} / on the \tjnum{27b}{jb_hard_rows}
  hard ones, under JevBench's scoring rule, with each model's release temperatures and thresholds
  (nothing fit on JevBench). Listed systems' outcomes are JevBench's own measurements at the pinned
  commit (Appendix~\ref{app:repro}). Exploratory. Source: \texttt{results/\allowbreak
  jevbench-<model>-report-v2.json}.}
  \label{tab:jevbench}
\end{table}

\section{Analysis and limitations}
\label{sec:limits}

\paragraph{Power.} The gate can only make a difference on decisions where System~1 and System~2
differ in correctness. For Qwen3.5-27B there are \tjnum{27b}{s1_wrong_s2_right} test decisions
where System~1 is wrong and System~2 right, and \tjnum{27b}{s1_right_s2_wrong} where thinking turns
a correct answer into a wrong one. Most fixes are in block~C (\tjnum{27b}{disagree_weak_fix} fixed,
\tjnum{27b}{disagree_weak_break} broken), then block~B (\tjnum{27b}{disagree_public_fix} and
\tjnum{27b}{disagree_public_break}); in block~A thinking fixes \tjnum{27b}{disagree_jev_fix}
answers but breaks \tjnum{27b}{disagree_jev_break}. Every model has at least
\tjnum{27b}{disagree_total} such test decisions (Qwen3.5-4B: \tjnum{4b}{disagree_total}), so the
comparisons between policies do not lack decisions where the choice matters.

\paragraph{One thought per decision.} System~2 is sampled once per decision, so its accuracy and every
anytime point carry sampling noise that a second seed would reduce. The equivalence between a
checkpoint and a thinking budget holds for the thought that was sampled, not for the expected
answer under that budget.

\paragraph{Data.} Public benchmarks may be in the model's training data; the generated block~C is
not, and we read the two side by side rather than pooling conclusions. The labels of the typed-decisions
dataset come from a single teacher model, not from people. Chinese decisions exist only in
block~C.

\paragraph{Calibration and thresholds.} Temperatures and thresholds are fit on a labeled dev split
from the same sources as the test decisions. They are not expected to transfer across domains,
and a deployment needs labeled decisions from its own domain. Nor does one temperature serve every
thinking budget: the raw logits grow more overconfident as the thought lengthens
(Section~\ref{sec:anytime}), so a deployment that changes its budget must refit. The risk
guarantee assumes that test groups are exchangeable with dev groups, \ie drawn from the same
distribution.

\paragraph{Scope of the readout.} Decisions have at most 16 options, each read through one
letter token, so free-form answers are out of scope. Rotations cover cyclic shifts of the listed
order, not arbitrary permutations.

\paragraph{Cost accounting.} Thinking, once triggered, is still slow: thought tokens dominate the
cost of an escalated decision. The anytime policy's cost counts thought tokens up to the checkpoint
where it stops, but not the short forward passes that read earlier checkpoints.

\paragraph{Comparisons we did not run.} We did not call Jev's API, so Jev is not evaluated on our
test set. The only comparison with Jev is on JevBench's public items (Section~\ref{sec:jevbench}), where
Jev's outcomes are JevBench's published measurements; other statements about Jev cite TypeSafe's own
pages. We also did not run DART~\citep{lee2026dart} or Sonata~\citep{li2026adaptive}; our probe
router stands in for the family of routers that read hidden states. The gate that
combines System~1's confidence with the zero-shot score (Section~\ref{sec:h1}) was fit post hoc,
after the planned analyses had been read, and only on System~1's correctness; a combination
trained on the gain from thinking, which needs System~2 labels on dev, was not run.

\section{Conclusion}
\label{sec:conclusion}

A typed decision can often be settled by a direct answer, but not when it takes several steps,
and thinking on every decision is slow. ThinkJev shows that one open hybrid-thinking model can
serve both cases: reading the same option letters with and without a thought, after one processed
prompt, gives a fast and a slow system whose probabilities are directly comparable, and reading
them part-way through one thought gives the answer under every thinking budget. No training is
needed.

Three findings matter for a deployment. First, direct answers can be used on their own with a
guaranteed error rate: at a 5\% target, Qwen3.5-27B answers \tjnum{27b}{s1_ltt_coverage_at_5pct}
of decisions from one forward pass, with an error rate of \tjnum{27b}{s1_ltt_test_risk_at_5pct} on
test, although smaller models can answer far fewer this way (Qwen3.5-4B: \tjnum{4b}{s1_ltt_coverage_at_5pct};
Qwen3.5-9B: \tjnum{9b}{s1_ltt_coverage_at_5pct}). Second, stopping a thought once its answer is
confident keeps most of thinking's accuracy (\tjnum{27b}{anytime_test_accuracy} against
\tjnum{27b}{always_s2_accuracy}) with \tjnum{27b}{anytime_thinking_share} of the thought tokens. Third,
System~1's confidence is a weak guide to when to think: our target of thinking on at most half of
the decisions failed on every model, because the decisions System~1 is least sure of are often ones
that thinking does not fix. On JevBench, an independent benchmark, the same settings put an open 27B
model in the accuracy band of the thinking APIs listed there.

The clearest next step is a router trained to predict where thinking helps, rather than where
System~1 is wrong, which needs System~2 labels on dev; sampling more than one thought per decision would
also reduce the noise in the anytime curve. Code, data generators, the IDs and splits of every decision,
evaluation reports, and a server for typed decisions are released at
\url{https://github.com/ding-xiaolu/ThinkJev}.

\bibliographystyle{plainnat}
\bibliography{refs}

\appendix
\section{The anytime readout equals a thinking budget}
\label{app:anytime}

Fix a decision, the thinking prompt's token IDs, the sampler (temperature, top-$p$, top-$k$), and
the decision's seed. Let $\theta$ be the thought this sampler emits when it may run to the budget $B$, and
$n = |\theta|$.

\paragraph{Claim.} For $t < n$, running the same sampler with budget $t$ emits $\theta_{1:t}$, and
closing it with the early-stop sentence $e$ and reading the answer slots gives the same logits as
the checkpoint read on $\theta_{1:t} \Vert e$. For $t \ge n$ the thought closes by itself before
the budget binds, so the answer under budget $t$ is the final System~2 read.

\paragraph{Conditions.} (i) Sampling is autoregressive and the random draw for position $i$
depends only on the seed and $i$, so the first $t$ tokens do not depend on the budget. (ii) The
early-stop sentence is appended only when the thought did not end by itself. (iii) The logits used
for sampling and for reading are numerically identical in both runs; with batched GPU kernels this
holds up to floating-point differences, which the tests bound (Appendix~\ref{app:repro}). In the
batch pipeline, $\theta$ comes from vLLM and the reads from a separate forward pass, so the
equivalence is with a budget-$t$ run of the vLLM sampler, and the read's agreement with the
single-process implementation is checked on replayed thoughts.

\paragraph{Implementation.} The thought is fed in chunks on one cache. At each checkpoint $t$ a
copy of the cache receives $e$ and the answer slots are read; the original continues. The final
System~2 read from the chunked cache matches a one-shot read, and each checkpoint read matches a
one-shot read of the cut thought (\texttt{tests/test\_pipeline\_tiny.py}).

\section{Generated decisions (block C)}
\label{app:data}

\texttt{benchmarks/generate\_owned.py} writes decisions whose labels are computed by code:
date arithmetic; conditional counting over a table; chained rules with exceptions and overrides;
stacked negation and quantifiers, with each answer checked against every possible case; multi-step arithmetic; logic grids whose
unique solution is checked by brute force; graph reachability and shortest paths; and scheduling
with overlap, capacity, and time zones. Each family has four difficulty levels that add reasoning
steps (more rows, conditions, rules, entities, hops, or a unit conversion) and is written natively
in English and in Chinese. Answer positions, answer classes, and for numeric options the answer's
rank among the values are balanced per family, language, and level; numeric distractors come from
typical slips. The default file has 2{,}400 English and 1{,}600 Chinese decisions; a test regenerates it
byte for byte and re-derives every label independently. The chess decisions turn Lichess puzzles
(CC0) with
popularity at least 80 and at least 1{,}000 plays into ``best move'' questions with four legal SAN
options, with difficulty from the rating bucket ($<$1200, 1200--1599, 1600--1999, $\ge$2000).
Table~\ref{tab:examples} shows the level-1 English decision of each family, abridged; the full
decisions are
\texttt{owned/<family>/en/d1/000} in the committed \texttt{benchmarks/data/owned-v1.jsonl}.

\begin{table}[h]
  \centering
  \small
  \begin{tabular}{@{}lp{0.62\linewidth}l@{}}
    \toprule
    Family & State and question (abridged) & Answer \\
    \midrule
    Dates & Notice given Thursday, June 24, 2027; 20 calendar days of notice, counted from the next
      day. On what date does the tenancy end? & July 14, 2027 \\
    Counting & An eight-row order log (city, status, channel, amount). How many orders have status
      `refunded'? & 1 order \\
    Chained rules & A purchasing policy with amount bands and a one-step reduction for pre-approved
      vendors, except hardware; a \$3{,}630 repair from a vendor not on the list. What approval is
      needed? & No approval \\
    Negation & ``It is false that crate A was not sealed.'' Was crate A sealed? & Yes \\
    Arithmetic & 12 full trays of 17 rolls plus 7 extra rolls; 110 sold. How many rolls are left? &
      101 rolls \\
    Logic grid & Three friends, three pets, three drinks, five clues. Who owns the turtle? & Jonas \\
    Graph & Six one-way roads between five places. Can you drive from Dunmore to Elm? & Yes \\
    Scheduling & Four bookings in one room; a meeting may start when another ends. Does 14:15--15:30
      overlap a booking? & Yes \\
    \bottomrule
  \end{tabular}
  \caption{One generated decision per family (level 1, English), abridged from the released file.}
  \label{tab:examples}
\end{table}

\section{Results per source}
\label{app:full}

Tables~\ref{tab:full-qwen35-27b}--\ref{tab:full-qwen38-27b} give, for every model, System~1, System~2,
and gate accuracy and the gate's escalation rate per source, language, and difficulty level. The gate
uses the threshold in \texttt{per\_family\_test\_gate}; each line is weighted by the number of
decisions over the families of \texttt{per\_family\_test}, and level ``--'' marks sources without
difficulty levels.

{\small
\begin{longtable}{@{}llllrrrrr@{}}
  \caption{Qwen3.5-27B: test accuracy and escalation per source, language, and difficulty level.}
  \label{tab:full-qwen35-27b} \\
  \toprule
  Block & Source & Lang. & Level & Rows & S1 & S2 & Gate & Escalated \\
  \midrule
  \endhead
  \tjtab{27b}{per_source}
  \bottomrule
\end{longtable}

\begin{longtable}{@{}llllrrrrr@{}}
  \caption{Qwen3.5-4B: test accuracy and escalation per source, language, and difficulty level.}
  \label{tab:full-qwen35-4b} \\
  \toprule
  Block & Source & Lang. & Level & Rows & S1 & S2 & Gate & Escalated \\
  \midrule
  \endhead
  \tjtab{4b}{per_source}
  \bottomrule
\end{longtable}

\begin{longtable}{@{}llllrrrrr@{}}
  \caption{Qwen3.5-9B: test accuracy and escalation per source, language, and difficulty level.}
  \label{tab:full-qwen35-9b} \\
  \toprule
  Block & Source & Lang. & Level & Rows & S1 & S2 & Gate & Escalated \\
  \midrule
  \endhead
  \tjtab{9b}{per_source}
  \bottomrule
\end{longtable}

\begin{longtable}{@{}llllrrrrr@{}}
  \caption{Qwen3.8-27B: test accuracy and escalation per source, language, and difficulty level.}
  \label{tab:full-qwen38-27b} \\
  \toprule
  Block & Source & Lang. & Level & Rows & S1 & S2 & Gate & Escalated \\
  \midrule
  \endhead
  \tjtab{q38}{per_source}
  \bottomrule
\end{longtable}
}

\section{Prompts}
\label{app:prompts}

Every read in this paper, on every source, uses the prompts below; nothing is tuned per benchmark.

\paragraph{System~1.} SemIf's \texttt{direct-options-v1}~\citep{semif2026}, unchanged. A decision
(state, question, options) becomes two chat messages. The system message is (one line in the
prompt):
\begin{verbatim}
Apply the supplied criterion to the supplied evidence. Choose exactly one
listed option. Respond with only its uppercase letter, with no explanation
or reasoning.
\end{verbatim}
The user message is one JSON object with the decision's state as \texttt{evidence}, its question as
\texttt{criterion}, and the options under the letters A, B, \dots\ in listed order. For the
request in Appendix~\ref{app:server}, the question \texttt{dept} becomes (one line in the prompt):
\begin{verbatim}
{"evidence": {"ticket": "I was charged twice for March."},
 "criterion": "Which team?",
 "options": [{"letter": "A", "description": "Payments"},
             {"letter": "B", "description": "tech"}]}
\end{verbatim}
An option without a description is listed under its ID. The messages are rendered with the
model's chat template with thinking disabled, which for Qwen3.5 ends the prompt with an empty
thinking block after the assistant's opening, and System~1 is the next-token distribution over the
letters at that position (Section~\ref{sec:s1}). Option IDs are never shown to the model, so the
rotations of Section~\ref{sec:s1} move descriptions between letters without other changes.

\paragraph{System~2.} The same two messages, rendered with thinking enabled, so the prompt ends
inside an open thinking block; the shared prefix is the longest common token prefix of the two
renderings (Section~\ref{sec:fork}). A thought that ends by itself is followed by the tokens that
follow the thinking block in the System~1 prompt. A thought cut at a checkpoint or at the budget is
closed with Qwen's early-stop sentence:
\begin{verbatim}
Considering the limited time by the user, I have to give the solution based
on the thinking directly now.
</think>
\end{verbatim}
(one line, preceded by a blank line and followed by one). The answer slots are then read at the
same letters as System~1's.

\paragraph{Zero-shot router.} The same System~1 prompt, applied to a new decision whose evidence is
the original decision (its state, question, and option descriptions, without letters), with the question
\begin{verbatim}
Before answering it, judge this decision task: can the criterion be decided
reliably at a glance, without step-by-step work such as calculation,
counting, date arithmetic, or chaining several rules?
\end{verbatim}
and two options, ``Yes: it can be decided reliably at a glance'' and ``No: it needs step-by-step
work before it can be decided''. Its score is the probability of the first.

\paragraph{Benchmark decisions.} Multiple-choice benchmarks (block~B) keep their question and options.
Classification sets list labels as options, within the 16-option limit: a BANKING77 decision offers 16
of the 20 most frequent intents (the gold one and 15 others chosen by hash), a CLINC150 decision the 15
intents of the gold intent's domain, so every distractor is a near miss; out-of-scope CLINC queries
form the separate escalation task. These subsets make the tasks differ from the usual full-label
protocols. JevBench items become decisions through the server's
mapping (Appendix~\ref{app:server}), so a JevBench choice item is read exactly like the example
above. \texttt{benchmarks/build\_cascade.py} builds every decision, and \path{tests/test_semif_parity.py}
pins the System~1 prompt.

\section{Server interface}
\label{app:server}

A request names a state and typed questions, as in TypeSafe's API~\citep{typesafe2026api}:
\begin{verbatim}
POST /v1/systemone
{"state": {"ticket": "I was charged twice for March."},
 "questions": {
   "dept": {"type": "choice", "instructions": "Which team?",
            "criteria": {"billing": "Payments", "tech": null}},
   "mood": {"type": "score", "instructions": "How angry?",
            "criteria": ["Calm", "Angry", "Furious"]}},
 "thinkjev": {"think": "auto", "include_thought": false}}
\end{verbatim}
A Choice keeps its options, a Noul becomes the options \texttt{true}/\texttt{false}, and a Score
becomes the levels $0, \dots, n-1$, whose value is the probability-weighted level. Each answer has
the shape of the corresponding TypeSafe answer type (choice or score with probabilities and
confidence; noul as a probability); confidence follows TypeSafe's published
formulas~\citep{typesafe2026confidence}, for a Choice with $n$ options $(p_{\max} - 1/n) / (1 - 1/n)$. The optional \texttt{thinkjev} field selects
\texttt{auto} (gated), \texttt{never} (System~1 only), or \texttt{always} (System~2), and the
response adds a \texttt{thinkjev} map with per-question traces: which system answered, System~1's
confidence, and the threshold. Differences from Jev: at most 16 options or levels per question,
questions run one after another, and \texttt{usage.output\_tokens} counts thought tokens. A test
round-trips a request through TypeSafe's SDK (\texttt{tests/test\_api.py}). With
\texttt{"stream": true} the same request answers as a server-sent event stream: System~1's
calibrated probabilities, then the answer read at each checkpoint as the thought passes it, so the
server's demo page draws Figure~\ref{fig:overview} for a live decision. ThinkJev is
independent of TypeSafe; only the public request format is shared.

\section{Reproduction}
\label{app:repro}

Models are pinned by Hugging Face revision:

\begin{center}
  \small
  \begin{tabular}{@{}ll@{}}
    Qwen3.5-4B & \texttt{851bf6e806efd8d0a36b00ddf55e13ccb7b8cd0a} \\
    Qwen3.5-9B & \texttt{c202236235762e1c871ad0ccb60c8ee5ba337b9a} \\
    Qwen3.5-27B & \texttt{fc05daec18b0a78c049392ed2e771dde82bdf654} \\
    Qwen3.8-27B & \texttt{1d4bf0f2ff6012fd82039f2fa52739d0dd7c60c0} \\
  \end{tabular}
\end{center}

\noindent SemIf's code is copied
from commit \texttt{1f2dea3}; a smoke run reproduces the \texttt{prompt\_sha256} of SemIf's
committed direct predictions, and \path{tests/test_semif_parity.py} pins the prompt. The
release split is \path{benchmarks/manifests/release-split.jsonl}; rebuilding it with
\texttt{build\_cascade.py assemble} must reproduce the file byte for byte. Commands for one model
end to end are in \path{scripts/run_model.sh}; every figure
and number in this paper is regenerated by
\begin{verbatim}
python benchmarks/make_figures.py --report results/<model>-b4096-report-v2.json \
    --label <Model> --out results/figures/<model> --regenerate \
    --manifest benchmarks/manifests/release-split.jsonl \
    --jevbench results/jevbench-<model>-report-v2.json
\end{verbatim}
(with \texttt{--latency} \path{results/qwen3.5-4b-latency.jsonl} for Qwen3.5-4B, and
\texttt{--example} \path{owned/dates/en/d2/050} for the Qwen3.5-27B decision in Figure~\ref{fig:overview}). The release reports are
\path{results/<model>-b4096-report.json} for \texttt{qwen3.5-4b}, \texttt{qwen3.5-9b},
\texttt{qwen3.5-27b}, and \texttt{qwen3.8-27b}, next to each model's per-decision cascade and router
outputs. They are attached to release \texttt{v0.1.0} of the repository as \path{results.tar.gz}, which
unpacks to \path{results/}, and \path{results/SHA256SUMS} lists the
SHA-256 of every report, cascade, router, and latency file
(\texttt{shasum -a 256 -c SHA256SUMS} in \path{results/} checks them). The
\path{results/<model>-b4096-report-v2.json} files, re-evaluated from the same cascade and router
outputs, add the post-hoc \texttt{combined} policy of Section~\ref{sec:h1} (a logistic regression on
the logits of \texttt{confidence\_rot} and \texttt{router\_zero\_shot}, fit on dev with five
group folds) and reproduce every other field of the release reports exactly; the figures and numbers
in this paper are generated from them.

\noindent The JevBench check (Section~\ref{sec:jevbench}) reads JevBench at commit\\
\texttt{7625ecb1b075cfe4d09d34141d353325825eea1d}: \path{benchmarks/export_jevbench.py} turns its
public items into decisions (IDs in \path{benchmarks/manifests/jevbench-public.jsonl}), the decisions run
through \path{scripts/run_model.sh} unchanged, and \path{benchmarks/evaluate_jevbench.py} writes
\path{results/jevbench-<model>-report-v2.json} from each model's release report and JevBench's
\path{results/v1.2/jevbench-v1.2-per-task.json}. As a pipeline check, JevBench's own
\texttt{typesafe} adapter was pointed at \texttt{thinkjev-serve} on five items, and the served
System~1 probabilities matched the batch run to within $10^{-6}$.

\end{document}